\documentclass{article}

\PassOptionsToPackage{numbers, compress}{natbib}
\usepackage[preprint]{neurips_data_2023}

\usepackage[utf8]{inputenc}     %
\usepackage[T1]{fontenc}        %
\usepackage{hyperref}           %
\usepackage{url}                %
\usepackage{booktabs}           %
\usepackage{amsfonts}           %
\usepackage{nicefrac}           %
\usepackage{microtype}          %
\usepackage{xcolor}             %
\usepackage[]{graphicx} %
\usepackage{multirow}
\usepackage{xr}

\newcommand{\eg}{\emph{e.g}\onedot}

\newcommand{\onedot}{.\null}

\newcommand{\Tref}[1]{Table~\ref{#1}}
\newcommand{\Eref}[1]{Equation~(\ref{#1})}

\newcommand{\Fref}[1]{Figure~\ref{#1}}

\makeatletter
\newcommand*{\addFileDependency}[1]{%
\typeout{(#1)}%
\@addtofilelist{#1}
\IfFileExists{#1}{}{\typeout{No file #1.}}
}\makeatother

\title{FineBio: A Fine-Grained Video Dataset of \\Biological Experiments with Hierarchical Annotation}

\author{
Takuma Yagi$^{1}$\hspace{1em}
Misaki Ohashi$^{2}$\hspace{1em}
Yifei Huang$^{2}$\hspace{1em}
Ryosuke Furuta$^{2}$\\
{\bf Shungo Adachi}$^{3}$\hspace{1em}
{\bf Toutai Mitsuyama}$^{1}$\hspace{1em}
{\bf Yoichi Sato}$^{2}$\\
$^1$Artificial Intelligence Research Center, \\National Institute of Advanced Industrial Science and Technology (AIST)\\
$^2$Institute of Industrial Science, The University of Tokyo\\
$^3$Department of Proteomics, National Cancer Center Research Institute\\
\texttt{\{takuma.yagi,mituyama-toutai\}@aist.go.jp} \\
\texttt{\{misaki-o,hyf,furuta,ysato\}@iis.u-tokyo.ac.jp}\\
\texttt{shadac2@ncc.go.jp}
}

\begin{document}

\maketitle

\begin{abstract}
In the development of science, accurate and reproducible documentation of the experimental process is crucial. Automatic recognition of the actions in experiments from videos would help experimenters by complementing the recording of experiments. Towards this goal, we propose FineBio, a new fine-grained video dataset of people performing biological experiments. The dataset consists of multi-view videos of 32 participants performing mock biological experiments with a total duration of 14.5 hours. One experiment forms a hierarchical structure, where a protocol consists of several steps, each further decomposed into a set of atomic operations. The uniqueness of biological experiments is that while they require strict adherence to steps described in each protocol, there is freedom in the order of atomic operations. We provide hierarchical annotation on protocols, steps, atomic operations, object locations, and their manipulation states, providing new challenges for structured activity understanding and hand-object interaction recognition. To find out challenges on activity understanding in biological experiments, we introduce baseline models and results on four different tasks, including (i) step segmentation, (ii) atomic operation detection (iii) object detection, and (iv) manipulated/affected object detection.
Dataset and code are available from \url{https://github.com/aistairc/FineBio}.
\end{abstract}

\section{Introduction}
In the development of science, accurate and reproducible documentation of the experimental process is crucial.
Biological experiments are a perfect example of this.
Reporting accurate materials, procedures, and results not only guarantees the correctness of the experiments but also becomes reliable evidence to promote new findings~\cite{howe2008future} and laboratory automation~\cite{holland2020automation}.
For desirable documentation of biological experiments, it is necessary to record what reagents are used, in what quantities, and by what operations, without omissions.
But if there are errors or oversights in keeping the record, it will be hard to reproduce the results.

\begin{figure}[t]
\centering
\includegraphics[width=1.0\linewidth]{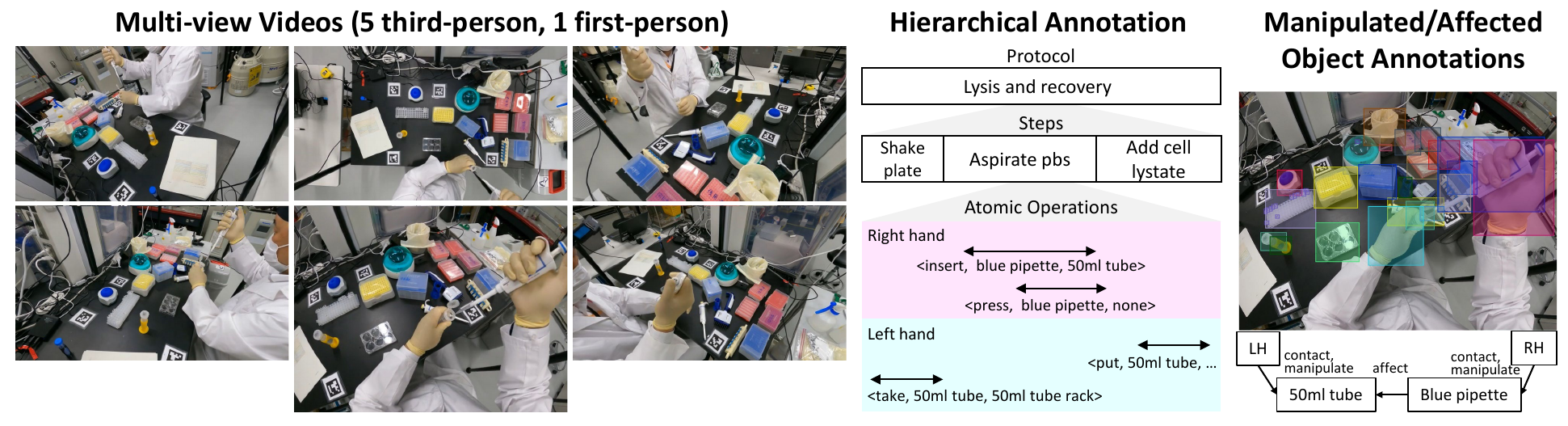}
\caption{We propose FineBio, a fine-grained and multi-view video dataset of biological experiments with hierarchical annotation. FineBio dataset consists of multi-view videos (left), hierarchical action annotation with different temporal granularity (middle), and frame-level object annotations (right).}
\label{fig:teaser}
\end{figure}

In this work, we aim to help researchers by automatically recognizing actions in experiments from video.
Specifically, we study the problem of recognizing various steps and operations in biological experiments from videos by collecting a new dataset of a person performing molecular biological experiments.
While few works collected footage of biological experiments~\cite{nishimura2021egocentric,grauman2022ego4d}, their annotations were only provided at the step level (\eg, add 1\,mL of PBS), and the amount of data was not enough to perform a quantitative evaluation by training models to recognize different steps and operations.
We propose FineBio, a controlled yet challenging set of biological experiment videos and annotation with multiple spatio-temporal granularity: steps, atomic operations, object bounding boxes, and their manipulation states (Figure~\ref{fig:teaser}).
We assume that one experiment follows a single protocol.
A protocol consists of several steps, and each step gives strict instructions to follow.
When performing one step, it is further decomposed into a set of atomic hand-related operations, such as ``inserting a pipette into a cell culture plate''.
Furthermore, such atomic operations occur from changes in hand-object relations (\eg, what object is manipulated by hand or influenced by other objects), which forms a hierarchical structure of an experiment.
Instead of focusing on one of the above, we aim to provide annotation at all levels to facilitate a holistic understanding of people performing experiments.
We record 226 trials totaling 14.5 hours and collect annotations of 3.5K steps, 50K atomic operations, and 72K object bounding boxes.
This enables us to train a model to recognize activities and objects.

To find out challenges in activity understanding in biological experiments, we introduce baseline models and their results in four tasks: (i) step segmentation (Section~\ref{ssec:step_segmentation}) (ii) atomic operation detection (Section~\ref{ssec:atomic_operation_detection}) (iii) object detection (Section~\ref{ssec:object_detection}), and (iv) manipulated/affected object detection (Section~\ref{ssec:manipulated_object_detection}).
The results show that the baseline models struggle in recognizing/detecting fine-grained details such as temporal boundaries and objects under interactions.

Our contributions are (i) a multi-view video dataset of 32 people performing mock biological experiments (ii) fine-grained action and object annotation on steps, atomic operations, object locations, and their manipulation states (iii) baseline models and experimental results on four different tasks, suggesting the need to model activity across multiple levels.

\section{Related Works}
\paragraph*{Video datasets on biological experiments}
While there are many image datasets that focus on part of biological experiments (\eg, \cite{edlund2021livecell,wei2021petri,lin2021nucmm,alam2019machine}), few works collect videos on performing biological experiments.
BioVL~\cite{nishimura2021egocentric} collects 16 egocentric videos of biological experiments with step-level caption annotation.
However, the number of videos is not enough to train a machine learning model and is only evaluated in a zero-shot setting.
Ego4D~\cite{grauman2022ego4d} contains 25 hours of videos performing biological experiments at a laboratory.
However, the activities are uncontrolled, unstructured, and only contain coarse narrations.
FineBio collects videos of specific molecular biological experiments, along with fine-grained hierarchical annotation, that enables training models across different granularity.

\paragraph*{Video datasets on structural activities}

Beyond recognizing individual actions, recognizing structured activities that consist of specific procedures is gaining more interest.
An activity can be divided into several {\it key steps}~\cite{alayrac2016unsupervised,bansal2022my}.
Recognizing such key steps from videos are studied in various fields such as cooking~\cite{kuehne2016end,papadopoulos2022learning}, sports~\cite{shao2020finegym,xu2022fine}, assembly~\cite{ragusa2021meccano,sener2022assembly101}, medical science~\cite{kaku2022strokerehab}, and ``how-to'' instructional videos~ including the aforesaid domains~\cite{alayrac2016unsupervised,tang2019coin,miech2019howto100m,Zhukov2019}.
A key step can be further divided into more fine-grained sub-actions.
Few datasets provide annotation on multi-level temporal action annotation~\cite{zhao2019hacs,shao2020finegym,rai2021home,sener2022assembly101}.
FineGym~\cite{shao2020finegym} is a dataset built on gymnastic videos with temporal annotation of three levels but does not incorporate interactions with the objects.
Assembly101~\cite{sener2022assembly101} offers a two-level hierarchy of coarse and fine-grained actions but differs in purpose in that the order of action is uncontrolled and left to participants, and lacks object-level annotation.
Our FineBio dataset offers comprehensive multi-level annotation from coarse yet strict step-level actions, sub-second atomic operations, and to frame-level object and manipulation state annotation in the practically important biology domain, providing a new challenge in recognizing actions and interacting objects across different temporal scales.

\paragraph{Hand-object relation modeling}
Predicting manipulated (active) objects and their manipulation states (\eg, physical contact) during hand manipulation is crucial in understanding hand-related activities.
Various tasks such as detecting object-in-contact~\cite{shan2020understanding}, contact prediction~\cite{narasimhaswamy2020detecting,yagi2021hand}, hand-object segmentation~\cite{shan2021cohesiv,darkhalil2022epic,zhang2022fine}, and active object detection~\cite{ragusa2021meccano,fu2022sequential} are studied.
However, most works focus on frame-level or short-term interactions, and annotation is not associated with action understanding tasks with a few exceptions~\cite{darkhalil2022epic}.
FineBio dataset provides object bounding box annotation with object-with-contact, manipulating objects, and affected objects, aligned with atomic operation annotation.
The trained hand-object relation prediction model could be used to improve action detection models and vice versa, which offers a unique value to the dataset.

\section{FineBio Dataset}
\label{sec:finebio}
\begin{table}[t]
\begin{center}
\begin{tabular}{cc}

\begin{minipage}{0.475\linewidth}
\begin{center}
\caption{Statistics on protocols.}
\scalebox{.75}{
\begin{tabular}{lrr}
\toprule
Protocol & \#Steps &\#Trials \\
\midrule
Lysis and recovery of cultured cells (1) & 9 & 46 \\
Lysis and recovery of cultured cells (2) & 12 & 45 \\
DNA extraction w/ magnetic beads (1) & 25 & 25 \\
DNA extraction w/ magnetic beads (2) & 31 & 26 \\
Polymerase chain reaction (PCR) & 10 & 46 \\
DNA extraction w/ spin columns (1) & 18 & 19 \\
DNA extraction w/ spin columns (2) & 21 & 20 \\
\bottomrule
\end{tabular}
}
\label{tab:recording_stats}
\end{center}
\end{minipage}

\begin{minipage}{0.475\linewidth}
\begin{center}
\caption{Splits statistics.}
\scalebox{.7}{
\begin{tabular}{lrrrrr}
\toprule
Splits & & Train & Valid & Test & Total \\
\midrule
\multicolumn{2}{r}{\# Participants} & 22 & 5 & 5 & 32 \\
\multicolumn{2}{r}{\# Trials} & 161 & 30 & 35 & 226 \\
\multicolumn{2}{r}{\# Steps} & 2,509 & 482 & 550 & 3,541 \\
\multicolumn{2}{r}{\# Atomic Operations} & 36,941 & 6,458 & 7,260 & 50,659 \\
\midrule
 & \# Frames & 1,394 & 238 & 303 & 1,935 \\
 & \# Bboxes & 50,886 & 8,867 & 11,795 & 71,548 \\
\bottomrule
\end{tabular}
}
\label{tab:finebio_split}
\end{center}
\end{minipage}

\end{tabular}
\end{center}
\end{table}

\begin{figure*}[t]
\centerline{\includegraphics[width=\linewidth]{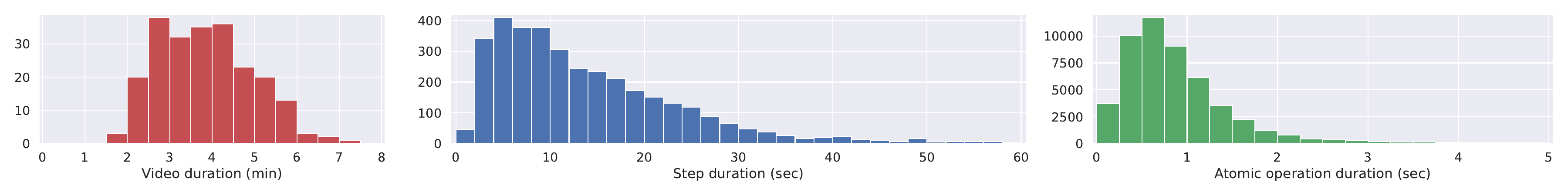}}
\caption{Distribution of duration: Average duration is 3.9~minutes, 14.3~seconds, 0.9~seconds, for protocols, steps, and atomic operations, respectively.}
\label{fig:duration_hist}
\end{figure*}

\begin{figure}[t]
\centerline{\includegraphics[width=\linewidth]{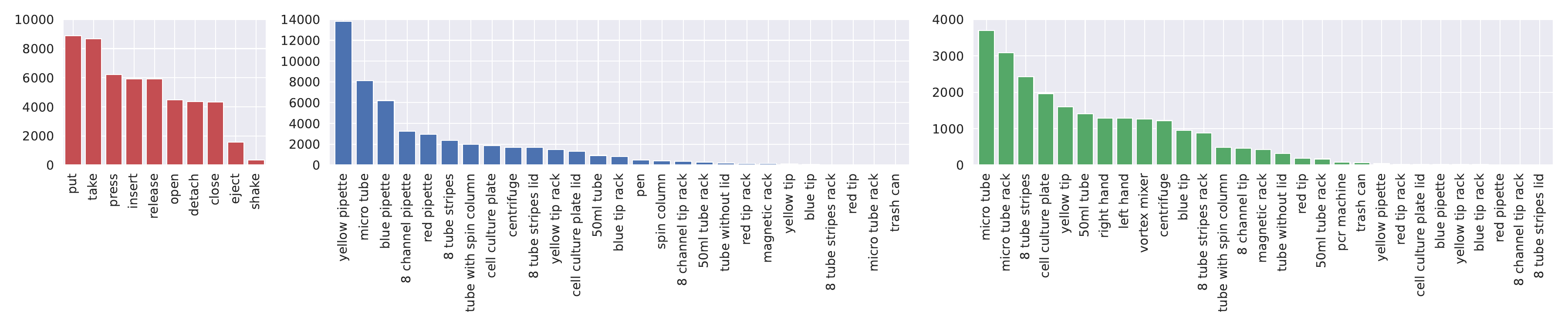}}
\caption{Distribution of verbs (left), manipulated objects (center), and affected objects (right).}
\label{fig:operation_hist}
\end{figure}

\subsection{Video Collection}
\label{ssec:collection}
\paragraph*{Protocols}
In this study, we define {\it protocol} as a set of instructions describing an experiment with a specific objective to be achieved.
FineBio consists of videos where a person is performing mock molecular biological experiments following specific protocols, in front of a table (see Figure~\ref{fig:teaser} left).
In actual biological experiments, special equipment (\eg, PCR machine, autoclave) is sometimes required and has waiting time till the chemical reaction process finishes.
Although these are crucial aspects in biological experiments, waiting time is tedious and special equipment may hinder the challenge of activity understanding by their device-specific operations.
Therefore, we modify some of the steps from actual protocols as follows: (a) use primary tools to focus on hand operations (b) use distilled water instead of real reagent (c) omit waiting time to centrifuge or purify molecules.

Specifically, we study four experiments on (i) lysis and recovery of cultured cells (ii) DNA extraction with magnetic beads (iii) polymerase chain reaction (PCR), and (iv) DNA extraction with spin columns, each taken from a popular procedure in amplification of DNA in cells.
From these experiments, we collect seven protocols consisting of multiple steps by changing the number of iterations of some steps (\eg, the iterations of ethanol wash).
Table~\ref{tab:recording_stats} shows statistics on the number of recordings.

\paragraph*{Camera configuration}
We use GoPro HERO9 Black for recording.
We record the experiments with five fixed third-person cameras ($4000\times3000$, 30~fps, linear FOV) and one head-mounted first-person camera ($3920\times2160$, 30~fps, wide FOV).
Each third-person camera was fixed to the platform and positioned left-back, left-front, right-back, right-front, and the above, of the participant (see Figure~\ref{fig:teaser} for example).
All cameras are temporally synchronized by a QR timecode displayed on a monitor ($<30~\mathrm{ms}$ error), and geometrically calibrated by a checkerboard for third-person cameras and four AR markers for first-person camera by following \cite{liu2021stereobj}.
Recording was conducted at two locations with a black working table.

\paragraph*{Participants and recording}
32 participants (16 men, 16 women) each performed 5--10 trials, resulting in 226 valid trials in total~\footnote{Four trials were excluded from evaluation due to critical procedural errors found after the recording, while another six trials with minor procedural errors were kept. See supplementary materials for detail.}.
The total recording duration was around 14.5 hours (87 hours of recording in total).
The average duration per trial was 3.9~minutes (see Figure~\ref{fig:duration_hist} left for details).
In contrast to \cite{sener2022assembly101}, we retook the trials if the participant made an unrecoverable mistake or performed steps in a wrong order.

\paragraph*{Objects}
To detect laboratory tools from a modest amount of annotation, we standardized the equipment used in the experiments.
33 pieces of equipment were selected to cover the objects used in the protocols (see supplementary materials for the full list).
Since the PCR machine was too large to be installed on the table, we substituted it with a tube rack with a distinctive appearance.

\subsection{Annotation}
\label{ssec:annotations}
We provide annotation at multiple granularitity: steps, atomic operations, object locations, and their states.

\paragraph{Step}
{\it Steps} refer to the ordered instructions described in a protocol.
The experimenter must strictly follow the instruction and the order of the steps to ensure reproducibility.
Therefore, the order of the steps is always the same in one protocol.
We define 32 categories for our dataset (see supplementary materials for detail).
Some steps are shared across different protocols.
We annotated the start/end time and the category of each step.
In total, 3,541 steps were annotated.
The average duration was 14.3~seconds, showing variation across categories (see Figure~\ref{fig:duration_hist} center for details).

\paragraph{Atomic operation}
{\it Atomic operation} is defined as a minimum meaningful hand-centric action in experiments.
Although steps are strictly regulated, how to manipulate the hands and equipment to accomplish the instructions in a step may differ across participants, and such operations may occur simultaneously.
Specifically, an atomic operation comprises a verb, manipulated object, and an affected object, each assigned to each hand.
{\it Verb} is defined by ten fine-grained motions such as put, take, insert, etc.
{\it Manipulated object} is defined as objects directly manipulated by the hand typically accompanied by contact between the hands.
We also define {\it affected object}, objects that are affected through the manipulated object (\eg, take a micro tube from a {\it tube rack}, insert a {\it tip} to a pipette).
Affected objects may not exist in some cases.
For example, collecting a sample in a micro tube to a pipette may be described in five operations: \{{\it (take, pipette, none)}, {\it(press, pipette, none)}, {\it (insert, pipette, micro tube)}, {\it (release, pipette, none)}, {\it (detach, pipette, micro tube)}\}.
In the case of using a conjunction of two or more objects (\eg, pipette and tip), we regard the manipulated object upon conjunction (pipette in this case) as a primary object.
We adopt Rubicon Boundaries~\cite{moltisanti2017trespassing} for determining the start and end time of the operations.
We collected 50,659 atomic operations in total.
The average duration per operation was 0.91~seconds.
As shown in Figure~\ref{fig:duration_hist} right, the majority of atomic operations were instant ($<1$\,seconds), making the operation detection more challenging.

\paragraph{Object location}
\begin{table}[t]
\begin{center}
\begin{tabular}{cc}
\begin{minipage}{0.475\linewidth}
\centerline{\includegraphics[width=\linewidth]{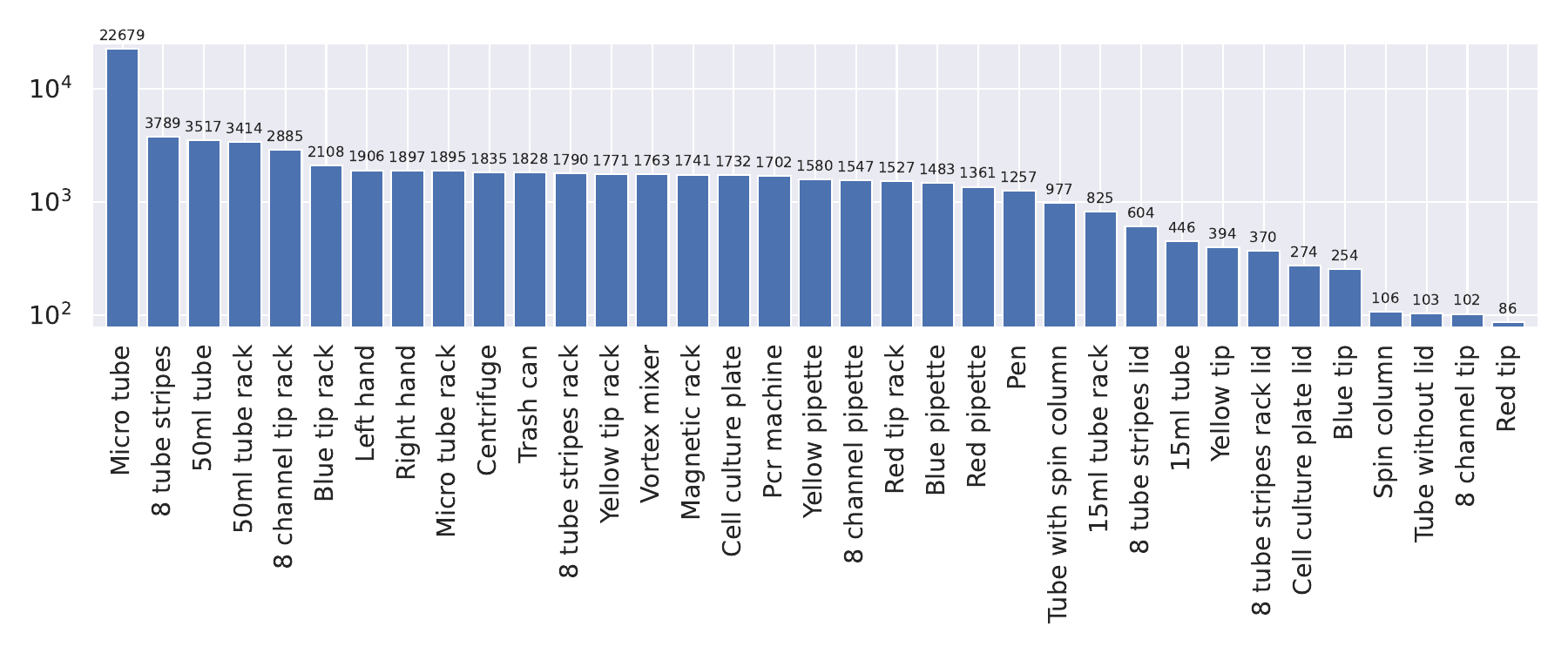}}
\caption{Distribution of object annotation (log scale).}
\label{fig:object_hist}
\end{minipage}

\begin{minipage}{0.475\linewidth}
\centerline{\includegraphics[width=\linewidth]{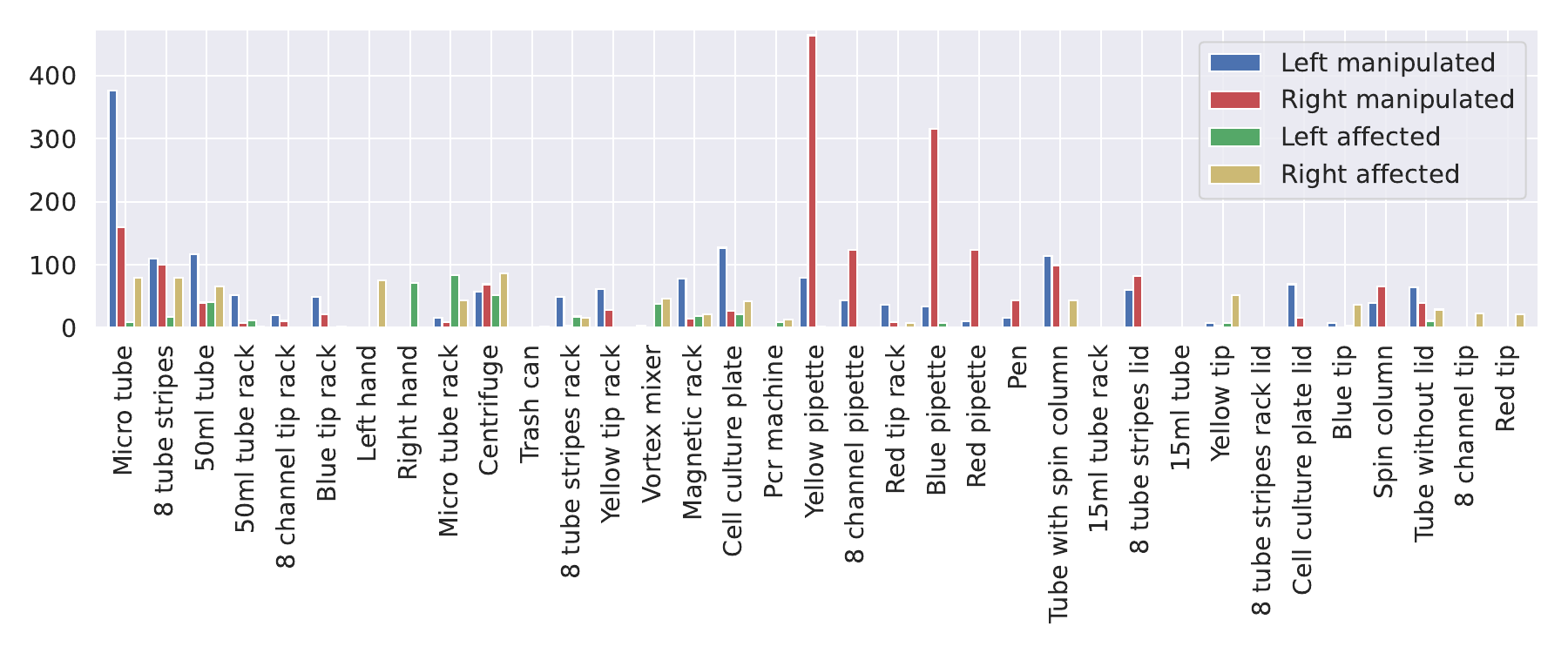}}
\caption{Distribution of manipulated/affected object annotation (linear scale).}
\label{fig:active_object_hist}
\end{minipage}
\end{tabular}
\end{center}
\end{table}

\begin{figure}[t]
\centerline{\includegraphics[width=\linewidth]{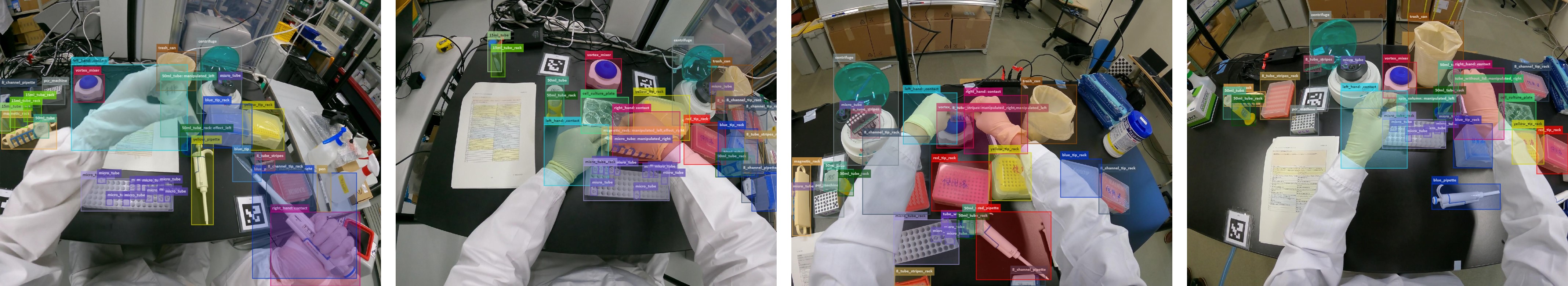}}
\caption{Examples of object bounding box annotation and their manipulation states. Each figure shows example of object annotation for protocol 1 (lyses and recovery), 3 (DNA extraction with magnetic beads), 5 (PCR), and 6 (DNA extraction with spin columns) from left to right. Each color of box denotes object category. Hand contact states and object manipulation states (contact, manipulated\_left/right, effect\_left/right) are shown next to the object name.}
\label{fig:object_annotation_example}
\end{figure}

To holistically understand activities, we also need to be able to determine which objects are involved in an operation.
Although several datasets~\cite{ragusa2021meccano,damen2022rescaling} annotate only {\it active objects} that are currently involved in the interactions due to cost constraints, they may harm performance of the object detector because other inactive objects are considered as background.
Besides active objects, {\it affected objects} that are not directly handled by hand but influenced through active objects are also important in analyzing activities but they were not considered in the above datasets.
In the above context, we decide to annotate all the hands and objects appearing in a frame rather than only active objects.
In addition, we annotate binary manipulation state information on the contact state, manipulated object, and affected object.

The contact state is assigned to each hand and is marked as true if an object has contact with the hand.
Similar to the definition of active objects as objects relevant to an action~\cite{damen2018scaling}, we define {\it manipulated object} as an object that is being manipulated within the interval of an atomic operation.
Similarly, we define {\it affected object} as an object that is being influenced within the interval of an atomic operation.
Therefore, some objects are considered manipulated even though they are not in contact (see Figure~\ref{fig:teaser} right).
Table~\ref{fig:object_hist} and \ref{fig:active_object_hist} show the number of annotated objects and manipulated/affected objects, respectively.

\paragraph{Selection of frames to annotate}
The appearance of an object may change significantly under interactions via motion and occlusion.
For a good object detector, collecting various samples that reflect such difficult scenarios is important.
To this end, we sample 1,935 frames from all the trials that contain novel situations (\eg, rare viewpoint, objects in contact, etc.) instead of uniform sampling.
Specifically, we first sample 1,346 first-person frames and further selected 118 frames from them to also annotate the corresponding 589 third-person frames (118 frames $\times$ 5 views, one omitted due to recording issue) that have the same timestamp as the first-person frames.
We annotated all the objects appearing in the frames, resulting in 71,548 bounding boxes in total.
We provide visual examples in Figure~\ref{fig:object_annotation_example}.

\begin{figure}[t]
\centerline{\includegraphics[width=\linewidth]{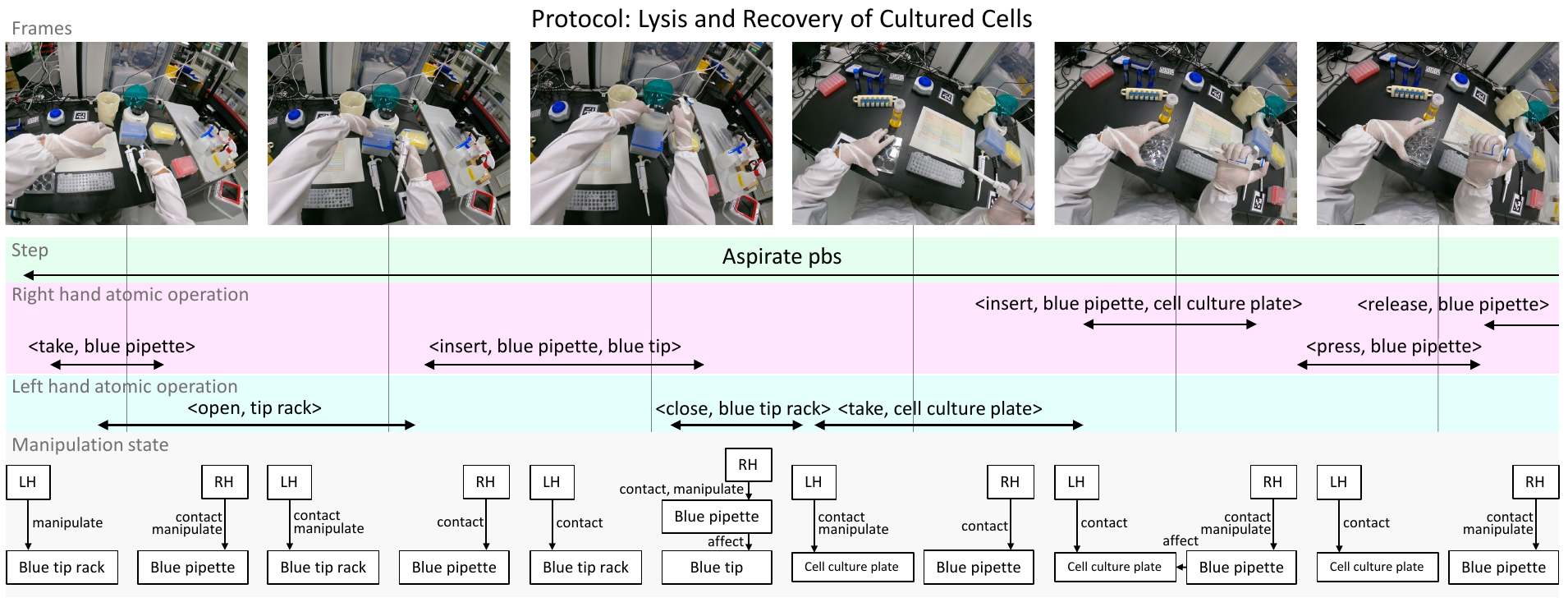}}
\caption{Example annotation of single step. LH and RH denote left hand and right hand, respectively. Note that ground truth contact annotation is only provided through object location annotation against sparsely sample frames.}
\label{fig:annotation_example}
\end{figure}

\paragraph{Hierarchical structure}
Figure~\ref{fig:annotation_example} shows example annotation of a specific step.
Each row shows the (first-person) video frame, task annotation, atomic operation annotation (right hand/left hand), and corresponding operation state, respectively.
As shown in the figure, atomic operations appear intermittently, sometimes with temporal overlap.
Although an experimenter can perform the operations at their convenient timing, one operation should satisfy their prerequisites.
For example, the tip rack must be opened before inserting a tip into a pipette.
A simple operation such as pushing a pipette will cause different results on whether it has been performed before or after inserting the pipette into a cell culture plate (collection or injection of the solution).
To recognize such small visual differences, not only looking at the holistic appearance but also capturing which objects are in-contact, manipulated, or affected will be important.
FineBio provides new challenges on how to develop a recognition model that can fuse information across different granularity.

\paragraph{Data splits}
Finally, we divided the data by participants, resulting in 22/5/5 participants for training, validation, and testing, without overlap.
Table~\ref{tab:finebio_split} summarizes the overall statistics of our FineBio dataset.

\section{Experiments}
\label{sec:experiments}
To find out challenges in holistic activity understanding in biological experiments, we introduce baseline models and report the results in four tasks: (i) step segmentation (Section~\ref{ssec:step_segmentation}) (ii) atomic operation detection (Section~\ref{ssec:atomic_operation_detection}) (iii) object detection (Section~\ref{ssec:object_detection}), and (iv) manipulated/affected object detection (Section~\ref{ssec:manipulated_object_detection}).
Each task corresponds to the annotation of each level, providing a set of challenging benchmarks on recognizing longer activities of tens of seconds to frame-level object states.
Although third-person videos and object annotation are also available, we only use first-person videos throughout this section.
All the code is available at \url{https://github.com/aistairc/FineBio}.

\subsection{Step Segmentation}
\label{ssec:step_segmentation}

\paragraph{Definition}
Given $T$ frames in a video, the goal is to predict step labels $\{s\}_{i=1}^T \subset S$ for all the frames. $S$ denotes a set of 32 step categories and a background label. 

\paragraph{Evaluation Metrics}
Following~\cite{li2020mstcn}, we report frame-wise accuracy, segmental edit score, and segmental F1 score with temporal overlapping thresholds at $k$\% (F1@$k$, $k\in\{10,25,50,75\}$).

\paragraph{Baselines and Results}
We use MS-TCN++~\cite{li2020mstcn} and ASFormer~\cite{yi2021asformer} as baselines for this task.
We extract features from video frames using I3D~\cite{carreira2017i3d} pretrained on Kinetics~\cite{carreira2017i3d} and input these features to the action segmentation models.
We fixed the feature extraction part and train the segmentation part from scratch using the training set.
Table~\ref{tab:results_seg} shows the quantitative results. 
Since the order of the steps are strictly defined by protocols and only trials with no ordering errors were included in the evaluation, the models showed high accuracy compared to unconstrained benchmarks~\cite{kuehne2016end,sener2022assembly101}.
Especially edit score and F1@\{10,25\} are over 90\%, which indicates both models can predict the presence and the category of the steps with high accuracy.

\begin{figure}[t]
\centerline{\includegraphics[width=\linewidth]{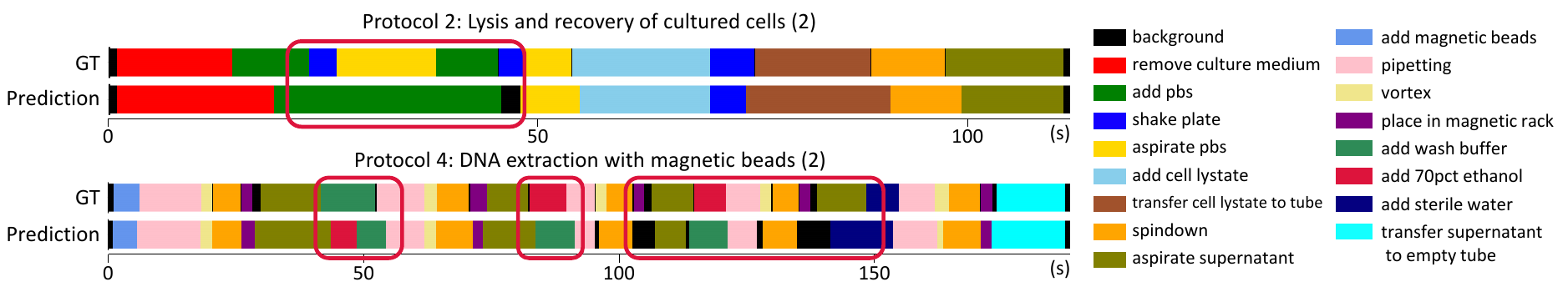}}
\caption{Failure cases in step segmentation. Red rectangles denote remarkable errors.}
\label{fig:qual_step_segmentation}
\end{figure}

\paragraph{Limitations and Challenges}
However, the score significantly drops at a stricter threshold (F1@75).
This suggests that the models struggles in inferring the exact boundary between steps at the operation level.
We show major failure cases of MS-TCN++ in Figure~\ref{fig:qual_step_segmentation}.
As shown inside the red rectangles, MS-TCN++ exhibited missing steps (upper), false positives, and totally wrong boundaries (bottom).
In the bottom example, ``add wash buffer'' and ``add 70pct ethanol'' were confused with each other since their operations within the steps are very similar (transferring different reagents from one tube to another).
To resolve the above issue, inferring the correspondence between a step and actions at a fine-grained operation level (\eg, making sure that the reagent has been surely loaded and delivered to the sample tube once) will be required.

\begin{table}[t]
\centering
\caption{Results on step segmentation.}
\small
\begin{tabular}{lrrrrrr}
\toprule
Model & Acc & Edit & F1@10 & F1@25 & F1@50 & F1@75  \\
\midrule
MS-TCN++~\cite{li2020mstcn} & 90.2 & 96.7 & 97.4 & 96.7 & 93.5 & 73.4 \\
ASFormer~\cite{yi2021asformer} & 87.2 & 94.8 & 94.2 & 92.7 & 86.5 & 67.0 \\
\bottomrule
\end{tabular}
\label{tab:results_seg}
\end{table}

\subsection{Atomic Operation Detection}
\label{ssec:atomic_operation_detection}

\paragraph{Definition}
The objective of this task is to localize the atomic operations and their constituents.
Adopting similar formulation to \cite{heilbron2015activitynet,damen2022rescaling}, given $T$ frames in a video, the goal is to predict a set of $N$ atomic operation instances $\mathbf{Y}=\{\mathbf{y}_i\}_{i=1}^N$, where $\mathbf{y}_i=(t_s, t_e, v, m, a)$ is a tuple consisting of start and end times $(t_s, t_e)$, verb $v$, manipulated object $m$, and affected object $a$.
Although atomic operations are assigned to each hand, this time we did not take the hand side into account and counted as duplicates if operations were done by both hands, for model simplicity.

\paragraph{Evaluation metrics}
We evaluate by the standard mean average precision (mAP) at different temporal IoU (tIoU) thresholds.
In addition to calculating the respective scores for verb, manipulated/affected objects, we calculate the mAP of {\it atomic operations} where counted as correct only if the predicted classes $(v, m, a)$ are all correct.

\paragraph{Baseline and results}
We choose ActionFormer~\cite{zhang2022actionformer} as an off-the-shelf method.
Input features are calculated from video frames using pre-trained I3D~\cite{carreira2017i3d} same as Section~\ref{ssec:step_segmentation}.
In the model, three classification heads are used to predict $v, m$, and $a$ respectively, followed by a single head taking concatenation of output probability vectors of the three heads as input, classifying the set $(v, m, a)$ from 244 possible combinations.
We used the same architecture as \cite{zhang2022actionformer} for each head.
The result is shown in Table~\ref{tab:results_det}. Each row shows mAP at different tIoU thresholds and their average score. We report scores by atomic operation, verb, manipulated object, and affected object, respectively. 
The results show that atomic operation detection is more challenging than step segmentation due to their unconstrained order.

\paragraph{Limitations and challenges}
The lower scores in object classes show difficulties in distinguishing similar-looking objects (\eg, blue and yellow pipette, tube with and without spin column) since the model does not explicitly detects objects.
See supplementary materials~\ref{sec:atomic_action_detection_details} for details.

\begin{table}[t]
\centering
\caption{Results on atomic operation detection by ActionFormer\cite{zhang2022actionformer}.}
\small
  \begin{tabular}{lrrrrrr}
  \toprule
    Target & mAP@0.3 & 0.4 & 0.5 & 0.6 & 0.7 & Avg. \\
    \midrule
    Atomic Operation & 45.2 & 41.7 & 36.5 & 28.4 & 18.7 & 34.1 \\
    \midrule
    Verb & 78.9 & 72.9 & 64.5 & 50.8 & 32.9 & 60.0 \\
    Manipulated Obj & 55.4 & 52.3 & 44.9 & 33.9 & 21.7 & 41.6 \\
    Affected Obj & 65.0 & 60.9 & 54.7 & 43.3 & 28.7 & 50.5 \\
    \bottomrule
  \end{tabular}
  \label{tab:results_det}
\end{table}

\subsection{Object Detection}
\label{ssec:object_detection}
\paragraph{Definition} 
The goal is to detect all objects' location and their categories (left/right hand + 33 tools) in a image.
Although detecting objects during manipulation is more important in a practical view, we first evaluate this standard-setting since we have access to exhaustive object location annotation.

\paragraph{Evaluation metrics}
We report COCO AP and Average Recall (AR)~\cite{lin2014microsoft}.
We report the AR of manipulated/affected objects to evaluate how well objects under interaction can be detected.

\paragraph{Baseline and results}
We adopt two transformer-based models (Deformable DETR~\cite{zhu2021deformable} and DINO~\cite{zhang2022dino}) as baseline models. We fine-tune the models pretrained on the MS COCO~\cite{lin2014microsoft} dataset. 
As shown in the results in Table~\ref{tab:results_object_det}, the overall scores were high in both models because the same sets of objects were used in training and testing.

\paragraph{Limitations and challenges.}
However, both models struggle with smaller objects, showing unsatisfying scores at $\mathrm{AP}_{\mathrm{S}}$ in particular.
Interestingly, we did not observe significant degradation in $\mathrm{AR}_{\mathrm{manip}}$ and $\mathrm{AR}_{\mathrm{affect}}$ compared to $\mathrm{AR}$, suggesting that producing bounding boxes of manipulated/affected objects under interaction itself is not a hard task.

\begin{table}[t]
\centering
\caption{Results on object detection.}
\small
  \begin{tabular}{lrrrrrrrrr}
  \toprule
    Method & $\mathrm{AP}$ & $\mathrm{AP}_{50}$ & $\mathrm{AP}_{\mathrm{S}}$ & $\mathrm{AP}_{\mathrm{M}}$ & $\mathrm{AP}_{\mathrm{L}}$ & $\mathrm{AR}$ & $\mathrm{AR}_{\mathrm{manip}}$ & $\mathrm{AR}_{\mathrm{affect}}$ \\
    \midrule
    Deformable DETR\cite{zhu2021deformable} & 53.3 & 77.4 & 11.1 & 30.8 & 59.8 & 62.1 & 55.9 & 51.6 \\
    DINO\cite{zhang2022dino} & 56.1 & 78.5 & 12.6 & 34.2 & 62.5 & 66.6 & 64.0 & 58.8 \\
    \bottomrule
  \end{tabular}
  \label{tab:results_object_det}
\end{table}

\subsection{Manipulated/Affected Object Detection}
\label{ssec:manipulated_object_detection}
To understand atomic operations primarily handled by hands, it is important to localize and recognize objects manipulated by hands.
It is also important to localize affected objects that are influenced through the manipulated object~\cite{yu2023fine} in the case of tool-use actions where a tool is used as a manipulated object to work on another object, {\it i.e.}, the affected object.
To this end, we extend the active object prediction scheme~\cite{shan2020understanding,darkhalil2022epic} and evaluate the performance on detecting manipulated and affected objects in video frames.

\paragraph{Definition}
Given single frame, the goal is to detect the hands and their manipulated objects as well as the affected objects.
Different from~\cite{shan2020understanding,darkhalil2022epic}, we also predict the object categories of the manipulated/affected objects.

\paragraph{Evaluation metrics}
We use COCO $\mathrm{AP}_{50}$~\cite{lin2014microsoft} following the prior works.
To evaluate the performance by elements, we calculate mAP for (1) left and right hands (2) manipulated objects of all object categories (3) affected objects of all object categories.
Then we report mAP of both hands, hands and manipulated objects both correct (H + Manipulated), and all three entities (H + M + Affected).

\paragraph{Baselines, results, and challenges}
We modify the Hand Object Detector~\cite{shan2020understanding} by adding additional branches.
First, we classify the hand side (left or right hand) and object categories at once in a single classification head. Our hand state classification head predicts if the hand manipulates something or not.
To detect one manipulated and one affected object, respectively, five branches are added for (a) binary manipulated state classification (manipulated or not) (b) manipulated object offset prediction (offsets from hand location, similar to~\cite{shan2020understanding}) (c) binary affecting state classification head (affecting or not) (d) binary affected state classification (afffected or not) (e) affected object offset prediction (offsets from manipulated object location).
We determine manipulated objects by first thresholding the detected objects by manipulation state prediction and associating to hands that are closest to hand location + manipulated object offset vector. Then we apply the same operations to the affected objects to associate the affected object and the manipulated object.
Different from prior works, we use all the annotated bounding boxes instead of using active object annotation solely.
We also replaced the base object detector to 4-scale DINO~\cite{zhang2022dino} with a ResNet-50 backbone.

\begin{figure}[t]
\centerline{\includegraphics[width=\linewidth]{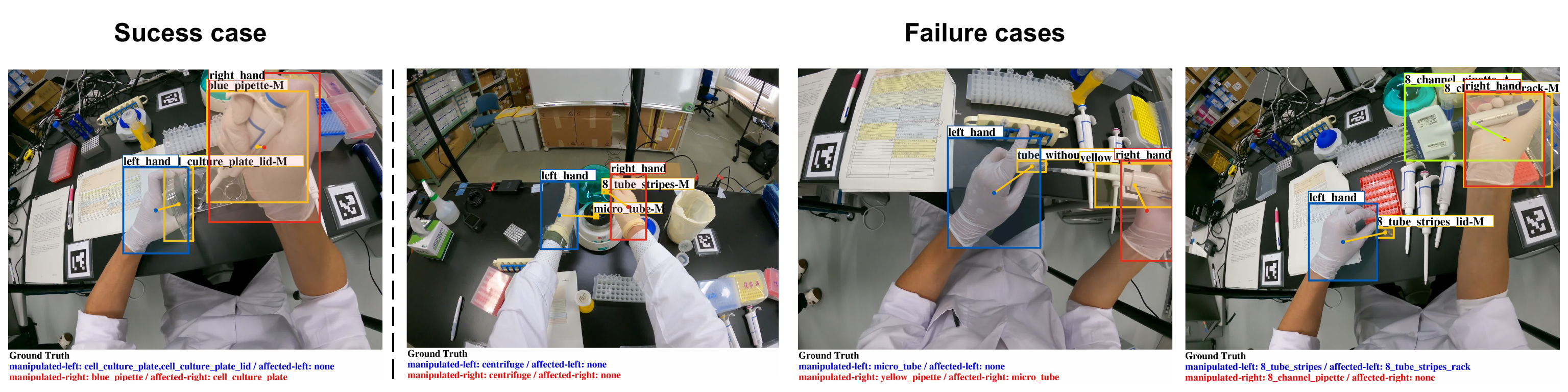}}
\caption{Qualitative results on manipulated and affected object detection. \textcolor{blue}{Blue}, \textcolor{red}{red}, \textcolor{yellow}{yellow} and \textcolor{green}{green} box denote left hand, right hand, manipulated object, and affected object. \textbf{M} and \textbf{A} next to the label name mean whether the object is manipulated or
affected object. The leftmost image shows successful case while the middle two display failures in manipulated object detection. Rightmost shows a failure in linking the correct affected object in affected object detection.}
\label{fig:qual_manip_object_det}
\end{figure}

We show the results in Table~\ref{tab:results_manip_object_det}.
Hands were almost perfectly detected.
However, detecting manipulated objects corresponding to hands was found to be very challenging despite the high object detection performance reported in Section~\ref{ssec:object_detection}.
Objects are densely arranged and some objects (\eg micro tube) are placed within other objects (\eg centrifuge), which makes it difficult to accurately identify the manipulated and affected objects.
The scores of manipulated/affected object detection by the left hand were especially low because smaller objects such as micro tube were often handled with the non-dominant left hand.
Figure~\ref{fig:qual_manip_object_det} shows the qualitative results.
While objects without occlusions were relatively easier, overlapping objects and smaller objects were hard to detect possibly due to the naive offset prediction scheme and lack of temporal context.

\begin{table}[t]
\centering
\caption{Results on manipulated/affected object detection (Box $\mathrm{AP}_{50}$).}
\small
  \begin{tabular}{lrrrr}
  \toprule
     & Hand & H + Manipulated & H + M + Affected \\
    \midrule
     Left hand & 96.8 & 6.5 & 5.9\\
     Right hand & 94.5 & 22.2 & 10.7 \\
    \bottomrule
  \end{tabular}
  \label{tab:results_manip_object_det}
\end{table}

\section{Conclusion}
\label{sec:conclusion}
We have presented FineBio, a fine-grained dataset of biological experiments with hierarchical annotation.
In addition to the extensive multi-level annotation, four benchmark tasks and their baseline evaluation have been presented.
We hope FineBio enhances the community to further develop new methods in the field of biological experiment understanding.

\paragraph{Limitations and future works}
For convenience and safety, we recorded ``mock'' experiments that are taken from real experiments but did not use real materials and reagents.
Some operations are simplified to remove redundant waiting time.
Therefore, the recognition model trained by this dataset cannot be directly applied to automating real biological experiments.
We focused on the operational aspects of experiments, such as whether or not the steps were conducted correctly in terms of protocols.
Thus we did not verify the experimental contents such as whether or not the amount of reagent is correct, whether or not the reaction occurred properly, and whether or not the goal is achieved.
Collecting experiments using real materials would be of interest as a future work.

\begin{ack}
Research at the University of Tokyo and National Institute of Advanced Industrial Science and Technology was supported by JST AIP Acceleration Research JPMJCR20U1.
\end{ack}

{\small
\def\nowcvpr{IEEE/CVF Computer Vision and Pattern
  Recognition}\def\pastcvpr{IEEE Computer Vision and Pattern
  Recognition}\def\iccv{IEEE/CVF International Conference on Computer
  Vision}\def\pasticcv{IEEE International Conference on Computer
  Vision}\def\tpami{IEEE Transactions on Pattern Analysis and Machine
  Intelligence}\def\bmvc{British Machine Vision
  Conference}\def\ijcv{International Journal of Computer
  Vision}\def\eccv{European Conference on Computer Vision}\def\wacv{IEEE/CVF
  Winter Conference on Applications of Computer Vision}\def\iclr{International
  Conference on Learning Representations}\def\neurips{Advances in neural
  information processing systems}\def\proc#1{Proceedings of the #1}

}

\section*{Checklist}

\begin{enumerate}
\item For all authors...
\begin{enumerate}
  \item Do the main claims made in the abstract and introduction accurately reflect the paper's contributions and scope?
    \answerYes{}
  \item Did you describe the limitations of your work?
    \answerYes{} See Section~\ref{sec:conclusion}.
  \item Did you discuss any potential negative societal impacts of your work?
    \answerYes{} See Section~\ref{sec:conclusion}.
  \item Have you read the ethics review guidelines and ensured that your paper conforms to them?
    \answerYes{}
\end{enumerate}

\item If you are including theoretical results...
\begin{enumerate}
  \item Did you state the full set of assumptions of all theoretical results?
    \answerNA{}
	\item Did you include complete proofs of all theoretical results?
    \answerNA{}
\end{enumerate}

\item If you ran experiments (e.g. for benchmarks)...
\begin{enumerate}
  \item Did you include the code, data, and instructions needed to reproduce the main experimental results (either in the supplemental material or as a URL)?
    \answerYes{} The code is publicaly available at \url{https://github.com/aistairc/FineBio}.
  \item Did you specify all the training details (e.g., data splits, hyperparameters, how they were chosen)?
    \answerYes{} Data splits are described at Section~\ref{ssec:annotations}. Training details of each task are described in Section~\ref{sec:experiments} and supplementary materials \ref{sec:step_segmentation_details}, \ref{sec:atomic_action_detection_details}, \ref{sec:object_detection_details}, and \ref{sec:manipulated_affected_object_detection_details}.
	\item Did you report error bars (e.g., with respect to the random seed after running experiments multiple times)?
    \answerNo{} We were not able to run all the experiments multiple times due to limited computation resources. We will publish the code to replicate the results.
	\item Did you include the total amount of compute and the type of resources used (e.g., type of GPUs, internal cluster, or cloud provider)?
    \answerYes{} See supplementary materials \ref{sec:resources}.
\end{enumerate}

\item If you are using existing assets (e.g., code, data, models) or curating/releasing new assets...
\begin{enumerate}
  \item If your work uses existing assets, did you cite the creators?
    \answerYes{} We cite the creators when using the code and models for baseline evaluation. See also supplementary materials \ref{sec:resources}.
  \item Did you mention the license of the assets?
    \answerYes{} The videos and annotation of the FineBio dataset will be available after agreeing to our original license agreement. The code will be available under the MIT license.
  \item Did you include any new assets either in the supplemental material or as a URL?
    \answerYes{} See supplementary materials datasheet \ref{sec:datasheet}.
  \item Did you discuss whether and how consent was obtained from people whose data you're using/curating?
    \answerYes{} See supplementary materials datasheet \ref{sec:datasheet}.
  \item Did you discuss whether the data you are using/curating contains personally identifiable information or offensive content?
    \answerYes{} See supplementary materials datasheet \ref{sec:datasheet}.
\end{enumerate}

\item If you used crowdsourcing or conducted research with human subjects...
\begin{enumerate}
  \item Did you include the full text of instructions given to participants and screenshots, if applicable?
    \answerYes{} We included the instructions and consent form translated from Japanese in the supplementary materials.
  \item Did you describe any potential participant risks, with links to Institutional Review Board (IRB) approvals, if applicable?
    \answerYes{} See supplementary materials datasheet \ref{sec:datasheet} for participant risks. Translated version of IRB approvals are also included in the supplementary materials.
  \item Did you include the estimated hourly wage paid to participants and the total amount spent on participant compensation?
    \answerYes{}  For the recording, participants consisted of 9 researchers from one of the authors' organizations (AIST) and 23 workers from a chemical industry company. Participants from both organizations participated in the recording as part of their duties. Although the exact amount of wages is unknown, when we outsourced the collection work to the latter company, we paid 1,600,000 JPY (approximately 70,000 JPY per participant). For the annotation, we have outsourced the annotation work to an agency and spent about 2,500,000 JPY in total.
\end{enumerate}

\end{enumerate}

\appendix
\section{Societal Impact and Resources Used}
\label{sec:resources}
\paragraph{Broader impacts}
We believe that the publicly released FineBio dataset and its corresponding benchmark suite promote AI development in understanding natural science experiments.
Specifically, video recognition techniques demonstrated in this study contribute to laboratory automation, improved reproducibility, and manipulation skill transfer.
Accurate recording of experiments will not only support researchers in performing experiments correctly but also prevent scientific misconduct such as fabrication and falsification.

\paragraph{Resources used}
We used a single NVIDIA A100 GPU for manipulated/affected detection, and a single NVIDIA Tesla V100 GPU for the other experiments. 
It took about an hour to train ActionFormer~\cite{zhang2022actionformer}. Step segmentation required 2.5-hour training for MS-TCN++~\cite{li2020mstcn} and 2-day training for ASFormer~\cite{yi2021asformer}. Training of Deformable DETR~\cite{zhu2021deformable} and DINO~\cite{zhang2022dino} took approximately 5.5 hours and 1 hour each. It took 15 minutes to train the hand-object detector for manipulated/affected detection.

\section{License of Assets}
\paragraph{License for code}
We used the official implementation of MS-TCN++~\cite{li2020mstcn,li2020mstcn_code}, ASFormer~\cite{yi2021asformer,zhang2022actionformer_code}, ActionFormer~\cite{zhang2022actionformer,zhang2022actionformer_code}, Hand Object Detector~\cite{shan2020understanding,shan2020understanding_code}, all of which are available under the MIT license. 
We also used the code of DINO~\cite{zhang2022dino_code}, Pytorch implementation~\cite{carreira2017i3d_code} of I3D~\cite{carreira2017i3d} and MMDetection~\cite{chen2019mmdetection}\cite{zchen2019mmdetection_code} under the Apache license, and RAFT~\cite{teed2020raft}\cite{teed2020raft_code} with permission by the BSD 3-Clause license.

\section{Dataset Additional Information}
\subsection{Data Release}
\label{ssec:release}

\paragraph{Documentation}
Our dataset and code are available at an official GitHub repository of Artificial Intelligence Research Center, National Institute of Advanced Industrial Science and Technology (\url{https://github.com/aistairc}).
The videos and annotation are hosted on ABCI Cloud Storage (\url{https://docs.abci.ai/en/abci-cloudstorage/}).
See Section \ref{sec:datasheet} for further details.

\paragraph{Data format}
All the data and annotations are released in standard formats.
Videos are provided in mp4 (h264 encoding) format.
Action (step and atomic operation) annotation is provided in CSV format, each row containing the start and end time, action, hand side (left or right), verb, manipulated object, and affected object.
Object annotations are provided in JSON format that extends the COCO format~\cite{lin2014microsoft}.
Precomputed feature vectors and camera pose information is provided in either text or numpy format.

\paragraph{Dataset split details}
For all the benchmark tasks, we split the dataset by participants as following:
\begin{itemize}
    \setlength{\leftskip}{-0.5cm}
    \item Training: P01, P02, P04, P06, P07, P10, P11, P12, P14, P16, P17, P18, P19, P21, P22, P23, P25, P26, P27, P29, P30, P31
    \item Validation: P05, P09, P15, P24, P32
    \item Testing: P03, P08, P13, P20, P28
\end{itemize}

\subsection{Details on Data Collection}
\begin{figure}[t]
\centerline{\includegraphics[width=\linewidth]{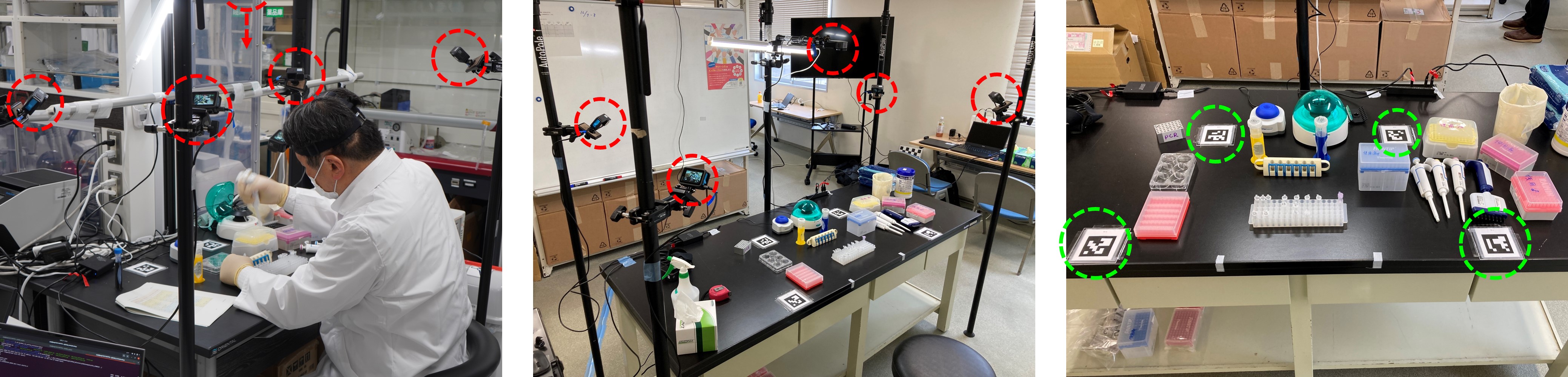}}
\caption{Camera and marker setup. (Left) Camera setup of the first environment (P01--P09). Red circles denote location of fixed third-person cameras. Top-view camera is out of frame. (Center) Camera setup of the second environment (P10--P32). (Right) Example of AR markers for extrinsic calibration of first-person camera. Green circles denote location of AR markers.}
\label{fig:camera_setup}
\end{figure}

\paragraph{Camera setup}
As stated in the main text, the five fixed cameras were positioned left-back, left-front, right-back, right-front, and above, of the participant.
Each camera was set up to view the entire surface of the desk (120cm $\times$ 60cm large for P01--P09, 160cm $\times$ 70cm large for P10--P32), and the wearer.
The top-view camera was positioned about 85--90~cm high from the desk surface.
Figure~\ref{fig:camera_setup} left and center shows the example setup of each environment.

\paragraph{Extrinsic calibration}
To obtain the extrinsic camera parameters of the third-person cameras, we resort to standard chessboard-based calibration~\cite{zhang2000flexible}.
For the first-person camera, we followed a similar procedure to StereOBJ-1M~\cite{liu2021stereobj}.
Specifically, we placed three or four AR markers on the table (see Figure~\ref{fig:camera_setup} right for example).
Then, we calculate the markers' position from the third-person cameras using Perspective-n-Point (PnP) pose computation.
After that, we calculate the first-person camera's location from the obtained marker position by triangulation.

However, in some cases, the camera positions were not perfectly obtained due to rapid head motion and severe occlusion by arms or equipment.
Therefore, the percentage of frames in which one or more AR markers were detected and the camera posture was recovered was 90.4\% on average.

\paragraph{Temporal synchronization}
\begin{figure}[t]
\centerline{\includegraphics[width=\linewidth]{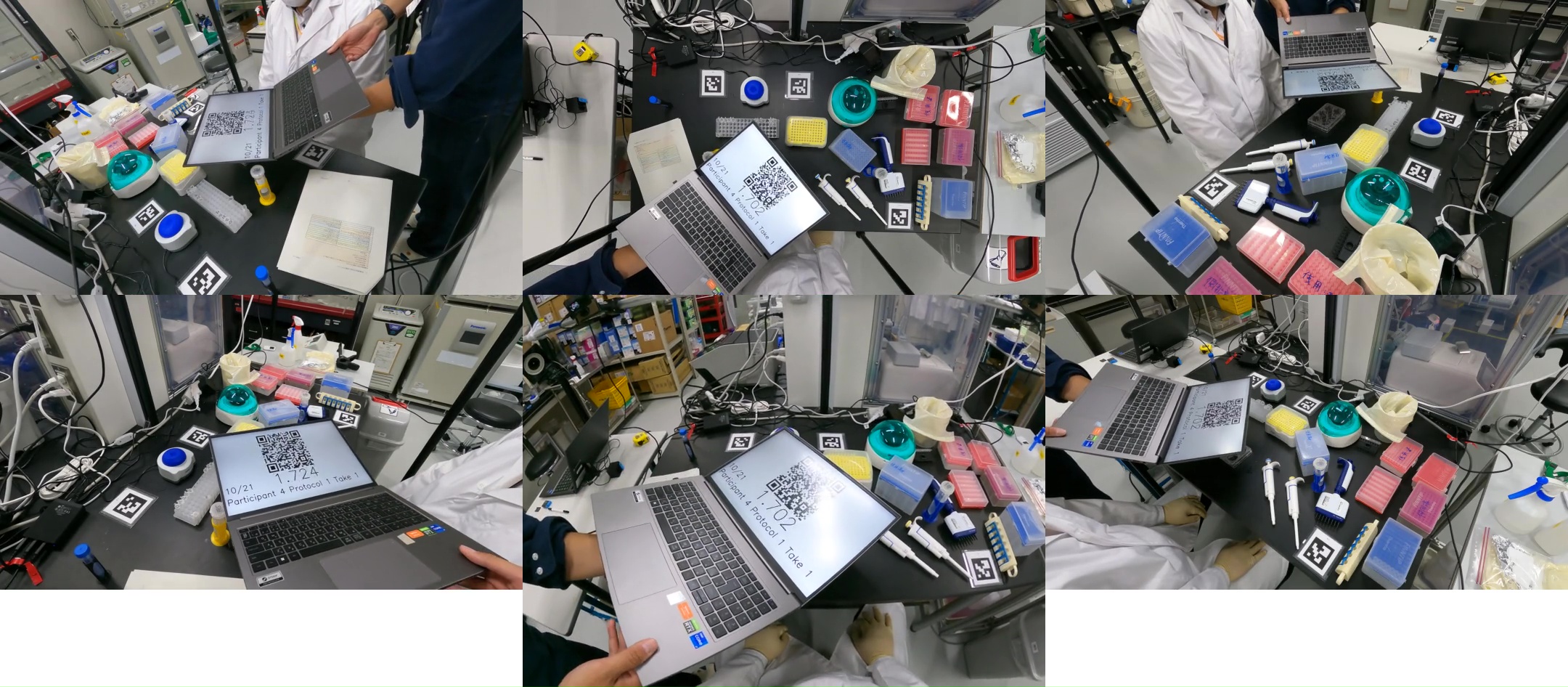}}
\caption{Examples of QR temporal synchronization.}
\label{fig:qr_sync}
\end{figure}

We implement an original QR code-based temporal synchronization scheme to synchronize videos across both third-person and first-person cameras.
At the beginning of each trial, we display a QR code that contains recording time, current elapsed time, participant id, protocol id, take id, etc. to efficiently manage the recordings.
Figure~\ref{fig:qr_sync} shows an example of a synchronization step.

\subsection{Details on Annotation}
\label{ssec:annotation_details}

\begin{table}[t]
\centering
\caption{Steps of protocol 1 and 2.}
\label{tab:protocol_1_2}
\scalebox{.9}{
\begin{tabular}{lll}
\hline
Step & Lysys and recovery of cultured cells (1) & Lysys and recovery of cultured cells (2) \\ \hline
1    & Remove culture medium                    & Remove culture medium                    \\
2    & Add 1mL of PBS                           & Add 1mL of PBS                           \\
3    & Gently shake the plate by hand           & Gently shake the plate by hand           \\
4    & Aspirate 1mL of PBS                      & Aspirate 1mL of PBS                      \\
5    & Add cell lystate                         & Add 1mL of PBS                           \\
6    & Gently shake the plate by hand           & Gently shake the plate by hand           \\
7    & Transfer the cell lysate to a 1.5mL tube & Aspirate 1mL of PBS                      \\
8    & Spindown                                 & Add cell lystate                         \\
9    & Aspirate supernatant                     & Gently shake the plate by hand           \\
10   &                                          & Transfer the cell lysate to a 1.5mL tube \\
11   &                                          & Spindown                                 \\
12   &                                          & Aspirate supernatant                     \\ \hline
\end{tabular}
}
\end{table}

\begin{table}[t]
\centering
\caption{Steps of protocol 3 and 4.}
\label{tab:protocol_3_4}
\scalebox{.9}{
\begin{tabular}{lll}
\hline
Step & DNA extraction with magnetic beads (1)           & DNA extraction with magnetic beads (2)           \\ \hline
1    & Dispense magnetic beads into sample tube         & Dispense magnetic beads into sample tube         \\
2    & Pipetting (when aspirating magnetic beads)       & Pipetting (when aspirating magnetic beads)       \\
3    & Pipetting (on dispensing)                        & Pipetting (on dispensing)                        \\
4    & Vortex                                           & Vortex                                           \\
5    & Spindown                                         & Spindown                                         \\
6    & Place micro tube in magnetic rack (for 1\,s) & Place micro tube in magnetic rack (for 1~s) \\
7    & Aspirate supernatant                             & Aspirate supernatant                             \\
8    & Dispense washing buffer into sample tube         & Dispense washing buffer into sample tube         \\
9    & Pipetting (on dispensing)                        & Pipetting (on dispensing)                        \\
10   & Vortex                                           & Vortex                                           \\
11   & Spindown                                         & Spindown                                         \\
12   & Place micro tube in magnetic rack (for 1~s) & Place micro tube in magnetic rack (for 1~s) \\
13   & Aspirate supernatant                             & Aspirate supernatant                             \\
14   & Dispense 70\% ethanol into sample tube           & Dispense 70\% ethanol into sample tube           \\
15   & Pipetting (on dispensing)                        & Pipetting (on dispensing)                        \\
16   & Vortex                                           & Vortex                                           \\
17   & Spindown                                         & Spindown                                         \\
18   & Place micro tube in magnetic rack (for 1~s) & Place micro tube in magnetic rack (for 1~s) \\
19   & Aspirate supernatant                             & Aspirate supernatant                             \\
20   & Dispense sterile water into sample tube          & Dispense 70\% ethanol into sample tube           \\
21   & Pipetting (on dispensing)                        & Pipetting (on dispensing)                        \\
22   & Vortex                                           & Vortex                                           \\
23   & Spindown                                         & Spindown                                         \\
24   & Place micro tube in magnetic rack (for 1~s) & Place micro tube in magnetic rack (for 1~s) \\
25   & Transfer supernatant to empty tube               & Aspirate supernatant                             \\
26   &                                                  & Dispense sterile water into sample tube          \\
27   &                                                  & Pipetting (on dispensing)                        \\
28   &                                                  & Vortex                                           \\
29   &                                                  & Spindown                                         \\
30   &                                                  & Place micro tube in magnetic rack (for 1~s) \\
31   &                                                  & Transfer supernatant to empty tube               \\ \hline
\end{tabular}
}
\end{table}

\begin{table}[t]
\centering
\caption{Steps of protocol 5.}
\label{tab:protocol_5}
\scalebox{.9}{
\begin{tabular}{lp{0.88\linewidth}}
\hline
Step & Polymerase chain reaction (PCR)                                                                                                       \\ \hline
1    & Dispense 20 {\textmu}L of sample from a 1.5mL tube into a 0.2mL 8-stripe tube\\
2    & Dispense reagent (PCR mix) from a 0.2mL 8-stripe tube into another 8-stripe tube containing a sample using an 8-channel pipette      \\
3    & Dispense reagent (forward primer) from a 0.2mL 8-stripe tube into the 8-stripe tube containing the sample using an 8-channel pipette \\
4    & Dispense reagent (reverse primer) from a 0.2mL 8-stripe tube into 8-stripe tube containing the sample using an 8-channel pipette     \\
5    & Dispense reagent (template dna) from a 0.2mL 8-stripe tube into 8-stripe tube containing the sample using an 8-channel pipette       \\
6    & Dispense water from a 0.2mL 8-stripe tube into 8-stripe tube containing the sample using an 8-channel pipette                        \\
7    & Cover the 8-stripe tube                                                                                                             \\
8    & Vortex the 8-stripe tube                                                                                                             \\
9    & Spin down the 8-stripe tube                                                                                                          \\
10   & Load the 8-stripe tube into the PCR machine                                                                                          \\ \hline
\end{tabular}
}
\end{table}

\begin{table}[t]
\centering
\caption{Steps of protocol 6 and 7.}
\label{tab:protocol_6_7}
\scalebox{.9}{
\begin{tabular}{lp{0.44\linewidth}p{0.44\linewidth}}
\hline
Step & DNA extraction with spin columns (1)                                    & DNA extraction with spin columns (2)                                    \\ \hline
1    & Dispense binding buffer into sample tube                                & Dispense binding buffer into sample tube                                \\
2    & Dispense ethanol solution into sample tube                              & Dispense ethanol solution into sample tube                              \\
3    & Vortex                                                                  & Vortex                                                                  \\
4    & Spindown                                                                & Spindown                                                                \\
5    & Transfer the solution from sample tube into tube with spin column       & Transfer the solution from sample tube into tube with spin column       \\
6    & Spindown                                                                & Spindown                                                                \\
7    & Dispense solution that has fallen to the bottom                         & Dispense solution that has fallen to the bottom                         \\
8    & Dispense washing buffer into sample tube                                & Dispense washing buffer into sample tube                                \\
9    & Spindown                                                                & Spindown                                                                \\
10   & Dispense solution that has fallen to the bottom                         & Dispense solution that has fallen to the bottom                         \\
11   & Dispense washing buffer into sample tube                                & Dispense washing buffer into sample tube                                \\
12   & Spindown                                                                & Spindown                                                                \\
13   & Dispense solution that has fallen to the bottom                         & Dispense solution that has fallen to the bottom                         \\
14   & Remove the spin column from the tubeand insert it into a new 1.5mL tube & Dispense washing buffer into sample tube                                \\
15   & Dispense the extract into the 1.5mL tube containing the spin column     & Spindown                                                                \\
16   & Spindown                                                                & Dispense solution that has fallen to the bottom                         \\
17   & Discard the spin column                                                 & Remove the spin column from the tubeand insert it into a new 1.5mL tube \\
18   &                                                                         & Dispense the extract into the 1.5mL tube containing the spin column     \\
19   &                                                                         & Spindown                                                                \\
20   &                                                                         & Discard the spin column                                                 \\ \hline
\end{tabular}
}
\end{table}

\paragraph{Details of protocols and steps}
Table~\ref{tab:protocol_1_2}, \ref{tab:protocol_3_4}, \ref{tab:protocol_5}, and \ref{tab:protocol_6_7} shows the procedure for each protocol.
The protocols are extracted from four experiments, where three of them have minor variations.
In {\it Lysys and recovery of cultured cells} (protocol 1, 2), the number of PBS washes was different (2 vs. 3 times).
In {\it DNA extraction with magnetic beads} (protocol 3, 4), the number of ethanol washes was different (1 vs. 2 times).
In {\it DNA extraction with spin columns} (protocol 6, 7), the number of centrifuges was different (2 vs. 3 times).
These minor variation makes the step recognition non-trivial because the steps will be non-unique across protocols, thus requiring the model to predict correct steps from the observed actions.

As explained in the main text, some of the time-consuming steps were intentionally modified to reduce the recording burden.
For example, the duration of vortex, spindown, and adsorption by magnetic beads was modified to a few seconds, which is not effective in real experiments.
Also, the end of protocol 5 (PCR) using the PCR machine was replaced by putting the 8-tube stripes on a silver tube rack since it does not include any meaningful hand operations.
We used distilled water instead of all reagents, thus it was not possible to observe appearance change due to chemical reactions.
Therefore, participants performed according to the protocol, imagining the actual experiment by themselves.

The beginning and end of a step are determined by when the first atomic operation conducted to achieve the goal of the step starts, and when the last atomic operation conducted to achieve the goal of the step ends, respectively.
Therefore, if there are undefined operations not directly related to the execution of experiments such as reading instructions between different steps, we marked them as background.

\paragraph{Atomic operation annotation}
\begin{figure}[t]
\centerline{\includegraphics[width=\linewidth]{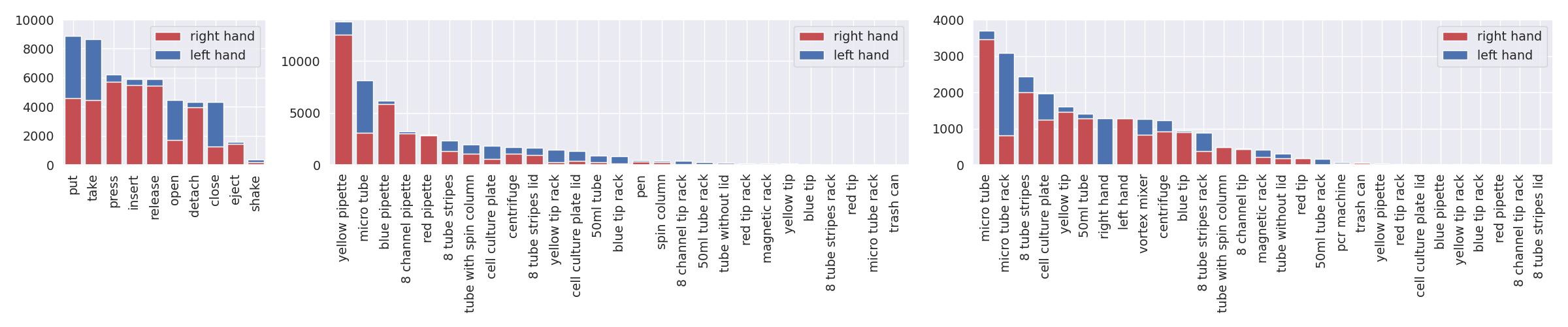}}
\caption{Distribution of verbs (left), manipulated objects (center), and affected objects (right) by hand side. Red and blue bars denote right hand and left hand action, respectively.}
\label{fig:operation_hist_by_hand}
\end{figure}
Figure~\ref{fig:operation_hist_by_hand} shows the occurance distribution of verbs, manipulated objects, and affected objects, each split by hand side.
Since most of the participants are right-handed, pipette-related actions are highly biased towards right hands, while few objects (\eg, micro tube, cell culture plate, tip rack) are manipulated by left hands.

\paragraph{Mistake annotation}

Although we intended to collect trials without procedural mistakes (\eg, violation of protocols, wrong step order), a small number of mistakes were identified after recording (see Table~\ref{tab:mistakes}).
Specifically, we found three major errors where multiple steps are missing or redundant.
Those trials were removed from the evaluation but kept in the released dataset for future purposes such as mistake detection.
In addition, we found eight minor errors that include single-step-level errors.
In this case, we determined the impact to be small and included these trials in the evaluation.

\begin{table}[t]
\centering
\caption{Identified mistakes in FineBio dataset.}
\scalebox{0.8}{
\begin{tabular}{lll}
\toprule
Video ID & Severity & Description of mistakes \\
\midrule
P06\_03\_02 & \multirow{3}{*}{Major} & Missing sterile water wash (6 steps missing) \\
P11\_06\_01 & & Extra wash buffer (3 steps redundant) \\
P17\_02\_02 & & Extra PBS wash (6 steps redundant) \\
\midrule
P03\_03\_02 & \multirow{8}{*}{Minor} & No pipetting at suction \\
P03\_04\_02 & & No pipetting at suction \\
P07\_04\_01 & & No vortex \\
P17\_07\_01 & & Mis-ordering of ``dispense solution'' and ``detach spin column and insert to new tube'' \\
P18\_02\_01 & & No spindown \\
P18\_05\_01 & & Last step missing due to missing frames \\
P24\_07\_01 & & Forget to dispense solution after second last spindown \\
P28\_06\_02 & & Mis-ordering of ``dispense solution'' and ``detach spin column and insert to new tube'' \\
\bottomrule
\end{tabular}
}
\label{tab:mistakes}
\end{table}

\paragraph{Object annotation}
The full list of 35 object categories (hands and equipment) are as follows:
left hand, right hand, blue pipette, yellow pipette, red pipette, 8-channel pipette, blue tip, yellow tip, red tip, 8-channel tip, blue tip rack, yellow tip rack, red tip rack, 8-channel tip rack, 50\,mL tube, 15\,mL tube, micro tube, 8-tube stripes, 8-tube stripes lid, 50\,mL tube rack, 15\,mL tube rack, micro tube rack, 8-tube stripes rack, 8-tube stripes rack lid, cell culture plate, cell culture plate lid, trash can, centrifuge, vortex mixer, magnetic rack, PCR machine, tube with spin column, spin column, tube without lid, pen.

We define object categories based on their intended usage rather than the general notion of an object. For example, tips inserted into an 8-channel pipette exist as a single unit like other tips, but in this specific protocol it is treated as an aggregate of eight tips, and thus the aggregate is treated as a single object.
The PCR machine is substituted by a silver tube rack instead of the actual equipment.
For the protocol {\it DNA extraction with spin columns}, the tube-like object {\it spin column} plays an important role in purifying nucleic acids, and is inserted into a micro tube to bind the nucleic acids while capturing other contaminants into the micro tube.
In this case, we decided to count the object ``tube with the spin column inserted'' as one combined object rather than treating it as two separate objects.

\subsection{Details on Conducting Annotation}
In this section, we briefly describe how the annotation was collected.
We collected the annotations through a crowdsourcing company (FastLabel Inc., \url{https://fastlabel.ai/}).
We spent 2,500,000 JPY in total for annotating step and atomic operation annotation of all 229 trials (14.5 hours) and all the 72K bounding boxes.
As FastLabel Inc. hired the workers as their duties, we could not report the hourly wage rate of the workers.

\paragraph{Quality control}
FastLabel offers quality control based on (i) instruction manual creation (ii) qualifier selection and training, and (iii) two-stage review.
First, we provide carefully-checked example annotation to FastLabel.
Then, FastLabel creates a detailed instruction manual for their workers and makes agreements on difficult cases through discussion.
Based on the instruction manual, FastLabel trains the workers and selects workers who are qualified to provide enough quality.
The selected workers provide annotation, and their results are examined by the inspector from FastLabel and finally by the authors.
To ensure accuracy from both a biological and machine learning perspective, experts from each field (TM and TY) reviewed all the trials before finalizing the annotation.

\paragraph{Annotation tools}
ELAN 6.4~\cite{elan} was used for step and atomic operation annotation.
The original object annotation tool provided by FastLabel was used for object location and manipulation state annotation.

\section{Additional Details on Step Segmentation}
\label{sec:step_segmentation_details}

\subsection{Implementation Details}
For feature extraction, we resize the longer side of each frame into 640 with their aspect ratio maintained.
We calculate optical flows by RAFT~\cite{teed2020raft} pretrained on Sintel dataset~\cite{butler2012sintel}.
We extract features from two-stream I3D model~\cite{carreira2017i3d} pretrained on Kinetics dataset~\cite{carreira2017i3d}.
We input a clip of 21 consecutive frames around each frame to produce a feature vector.

For training of MS-TCN++~\cite{li2020mstcn}, we set the number of layers to 12 in the prediction stage, 13 in the refinement stage, and the number of refinement stages to four. 

For training of ASFormer~\cite{yi2021asformer}, we set the number of blocks per encoder/decoder to 11 and the number of the decoders to five. 

We train both models with an initial learning rate of $5\times10^{-4}$ and batch size of 1. We use Adam~\cite{kingma2014adam} optimizer with the cosine annealing scheduler for MS-TCN++ and the ReduceLROnPlateau scheduler for ASFormer. We set the number of epochs to 100 for MS-TCN++ and 120 for ASFormer. We use the training, validation, and test split for training, hyperparameter tuning, and evaluation, respectively.

\section{Additional Details on Atomic Operation Detection}
\label{sec:atomic_action_detection_details}
\subsection{Implementation Details}
Similar to step segmentation, we resize the longer side of each frame into 640 with their aspect ratio maintained. We calculate optical flows by RAFT~\cite{teed2020raft} pretrained on Sintel dataset~\cite{butler2012sintel}. We extract features from two-stream I3D model~\cite{carreira2017i3d} pretrained on Kinetics dataset~\cite{carreira2017i3d}.
Finally, we input a clip of 16 consecutive frames at a stride of 4 to calculate the feature vector.

On training ActionFormer~\cite{zhang2022actionformer}, we use two convolution layers for projection, seven transformer blocks for the encoder with 2$\times$ downsampling for the last five blocks. The number of pyramid feature levels is six and a regression range on $i$-th level is [$2^{i-1}$, $2^{i+1}$). We set a window size of nine for local self-attention. We train the model with an initial learning rate of $10^{-4}$ and batch size of 1 for 40 epochs with a linear warmup of 5 epochs. We use the Adam optimizer with the cosine annealing scheduler. We use the training, validation, and  test split for training, validation, and evaluation, respectively. 

\subsection{Ablation on Different Model Architectures}
At each timestep, Actionformer predicts the action label and the distance to the onset and offset of the action by using a classification head and a regression head.
For atomic operation detection, We examine four types of model architectures with different head structures. 
{\bf Set classification}: This architecture has a single classification head to classify the atomic operation as a set from 224 all possible combinations.
{\bf Separate heads}: This architecture has three heads each predicting the verb, manipulated object, and affected object, separately.
{\bf Separate heads + Set classification}: This architecture first predicts each entity by three classification heads, followed by an additional classification head to classify the atomic operation as a set from 224 all possible combinations.
{\bf Separate models}: This architecture predicts the three entities with three separate models without weight sharing.
Both the classification heads and the regression heads are separately trained for each entity.

For {\bf Separate heads} and {\bf Separate models}, we calculate the classification scores of atomic operations by using \Eref{eq:action_detection_combine_scores} when evaluating APs on atomic operation $(v, m, a)$:
\begin{equation}\label{eq:action_detection_combine_scores}
    \mathbf{p}(v, m, a) = \mathbf{p}(v)^\alpha \times \mathbf{p}(m)^\beta \times \mathbf{p}(a)^{(1-\alpha-\beta)}.
\end{equation}
$v, m, a$ indicate verb, manipulated object, and affected object, respectively.
$\mathbf{p}$ denotes the classification score of each entity after each classification head.

Because {\bf Separate models} have separate regression heads, we use \Eref{eq:action_detection_combine_offsets} to calculate the weighted average of the distance to the start or the end time of each entity to predict the temporal boundary of atomic operations:
\begin{equation}\label{eq:action_detection_combine_offsets}
    d_{(v, m, a)} = \alpha d_{v} + \beta d_{m} + (1-\alpha-\beta) d_{a},
\end{equation}
where $d_{v}, d_{m},d_{a}$ denote the distance calculated in the regression head for each entity and $\alpha, \beta$ are weight coefficients.
We set $\alpha, \beta$ to $1/3$ each.

\Tref{tab:action_detection_compare_arc} shows the ablation results.
The models with separate heads performed better when evaluated separately.
However, for the atomic operation prediction models with set classification outperformed the separate prediction models suggesting the need of modeling relationships between entities.
Having separate heads for all four entities ({\bf Separate heads + Set classification}) performed the best overall.

\begin{table}[t]
\centering
\caption{Results on atomic operation detection by ActionFormer~\cite{zhang2022actionformer} with different architectures.}
\small
  \begin{tabular}{lrrrrrrr}
  \toprule
    Target & Architecture & mAP@0.3 & 0.5 & 0.7 & Avg. \\
    \midrule
        \multirow{4}{*}{Atomic operation} & Set classification & 42.9 & 33.4 & 17.2 & 31.9 \\
    & Separate heads & 37.5 & 30.1 & 15.9 & 28.3 \\
    & Separate heads + Set classification & {\bf 45.2} & {\bf 36.5} & {\bf 18.7} & {\bf 34.1} \\
    & Separate models & 36.8 & 29.6 & 16.2 & 28.1 \\
    \midrule
    \multirow{4}{*}{Verb} & Single head & 73.3 & 59.3 & 30.7 & 55.7 \\
    & Separate heads & 77.6 & 63.0 & 32.2 & 58.5 \\
    & Separate heads + Set classification & {\bf 78.9} & {\bf 64.5} & {\bf 32.9} & {\bf 60.0} \\
    & Separate models & 77.9 & 63.5 & 31.6 & 59.1 \\
     \midrule
    \multirow{4}{*}{Manipulated object} & Set classification & 54.7 & 43.8 & 21.9 & 41.0 \\
    & Separate heads & {\bf 56.0} & {\bf 45.6} & {\bf 21.8} & {\bf 41.9} \\
    & Separate heads + Set classification & 55.4 & 44.9 & 21.7 & 41.6 \\
    & Separate models & 53.3 & 41.7 & 20.0 & 39.2 \\
     \midrule
    \multirow{4}{*}{Affected object} & Set classification & 62.5 & 51.1 & 26.3 & 48.7 \\
    & Separate heads & {\bf 65.2} & 54.1 & 27.6 & 49.9 \\
    & Separate heads + Set classification & 65.0 & {\bf 54.7} & {\bf 28.7} & {\bf 50.5} \\
    & Separate models & 64.2 & 51.8 & 26.8 & 48.5 \\
    \bottomrule
  \end{tabular}
  \label{tab:action_detection_compare_arc}
\end{table}

\subsection{Error Analysis}
\begin{figure}[t]
    \begin{minipage}[htbp]{0.45\linewidth}
        \centering
        \includegraphics[scale=0.2]{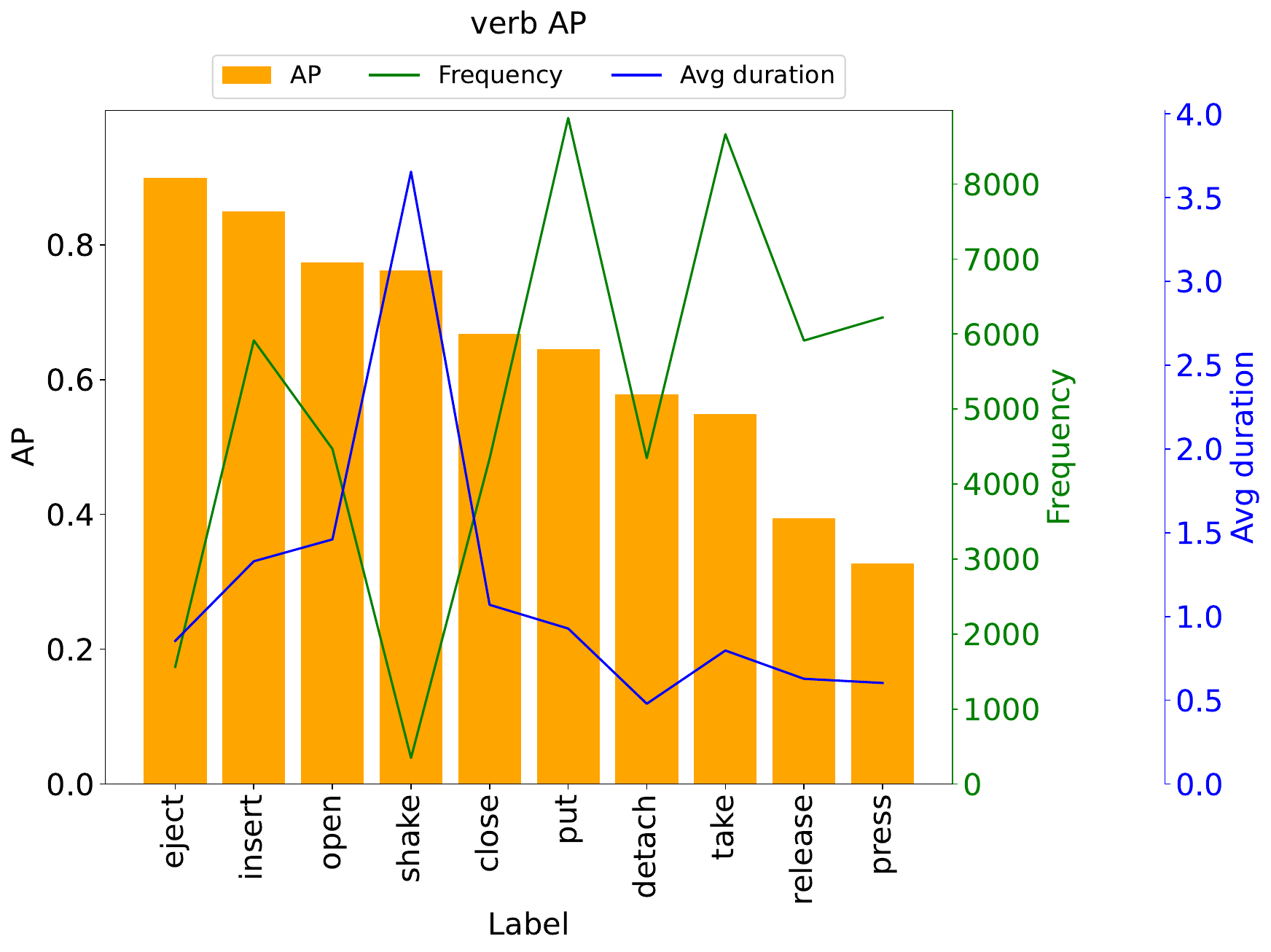}
    \end{minipage}
    \begin{minipage}[htbp]{0.45\linewidth}
        \centering
        \includegraphics[scale=0.2]{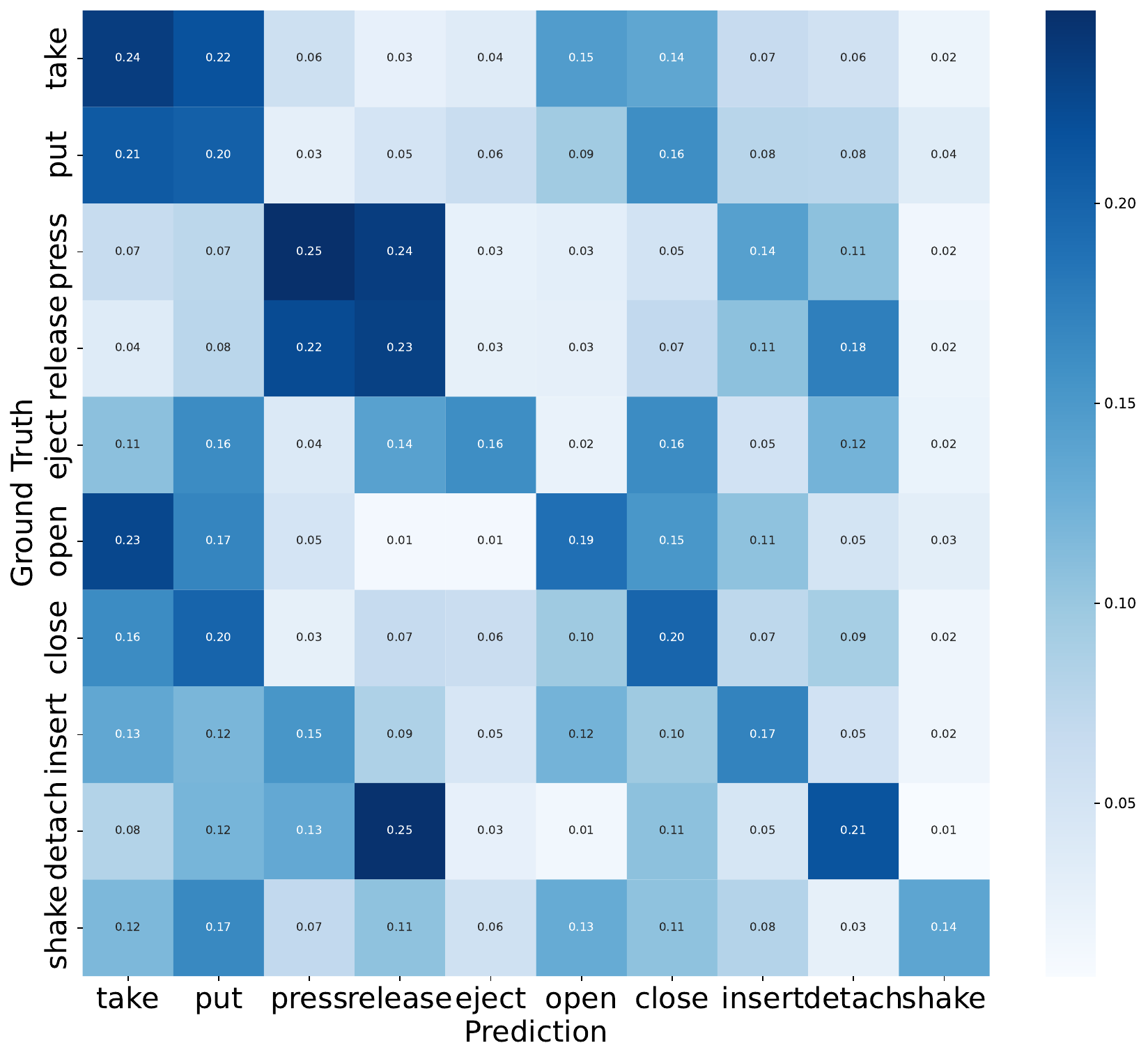}
    \end{minipage}
    \caption{Average Precision (left) and confusion matrix (right) for verb classes. The green and blue lines in the left figure denote average duration (in seconds) and number of occurrence of each verb.}
    \label{fig:verb_ap_confmat}
\end{figure}

\begin{figure}[t]
    \begin{minipage}[htbp]{0.45\linewidth}
        \centering
        \includegraphics[scale=0.2]{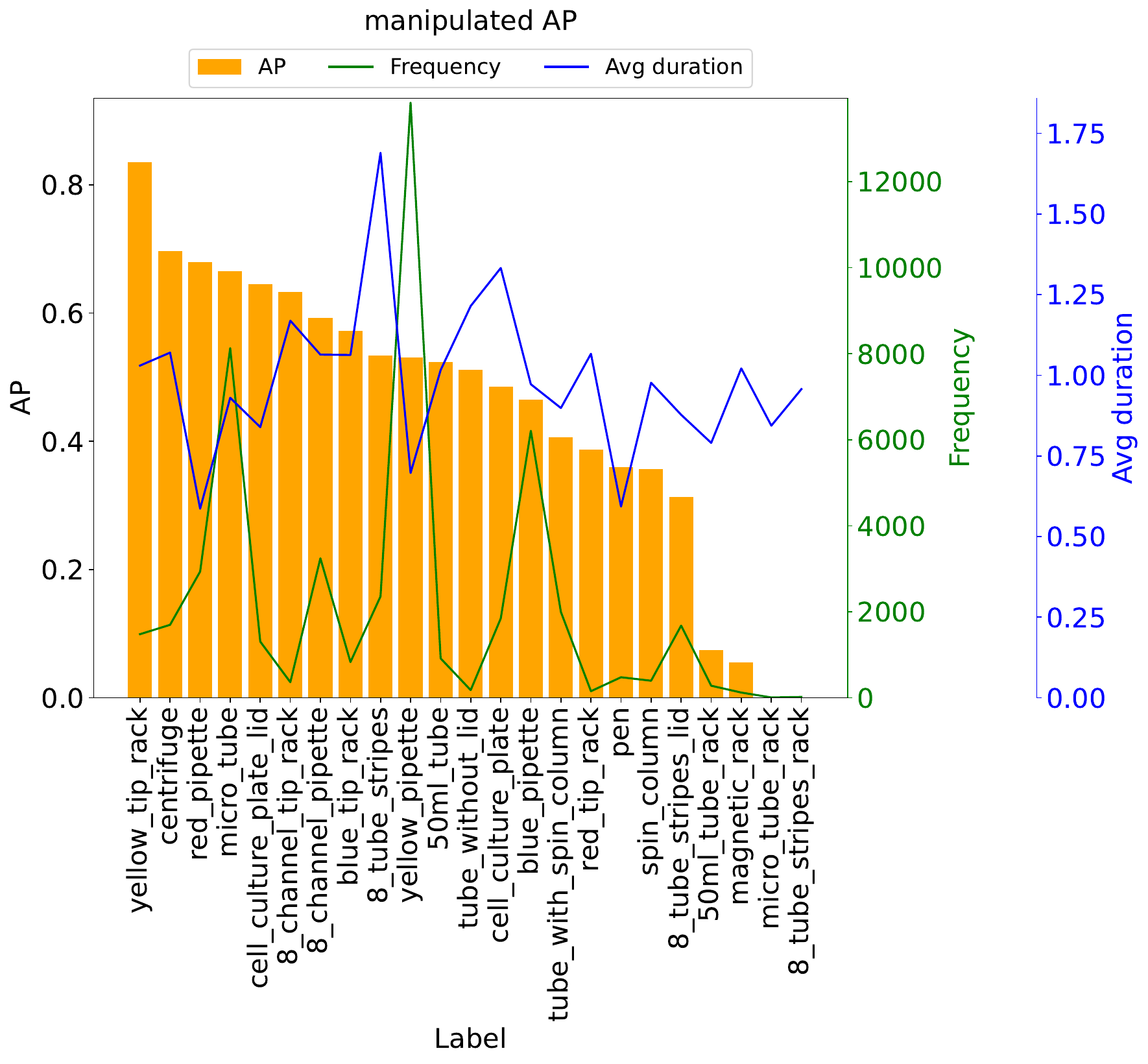}
    \end{minipage}
    \begin{minipage}[htbp]{0.45\linewidth}
        \centering
        \includegraphics[scale=0.16]{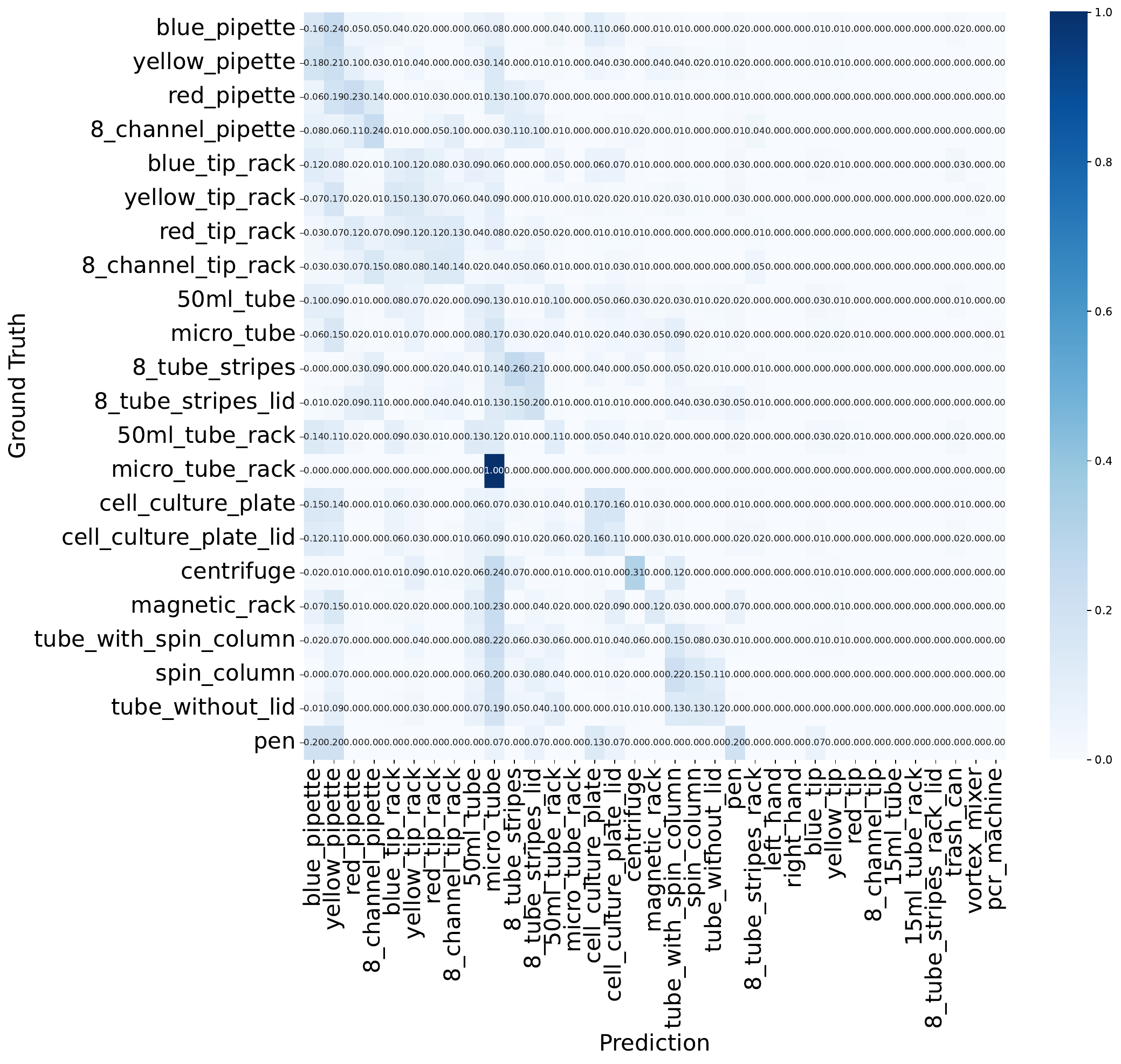}
    \end{minipage}
    \caption{Average Precision (left) and confusion matrix (right) for manipulated object classes. The green and blue lines in the left figure denote average duration (in seconds) and number of occurrence of each verb.}
    \label{fig:manip_ap_confmat}
\end{figure}

\begin{figure}[t]
    \begin{minipage}[htbp]{0.45\linewidth}
        \centering
        \includegraphics[scale=0.2]{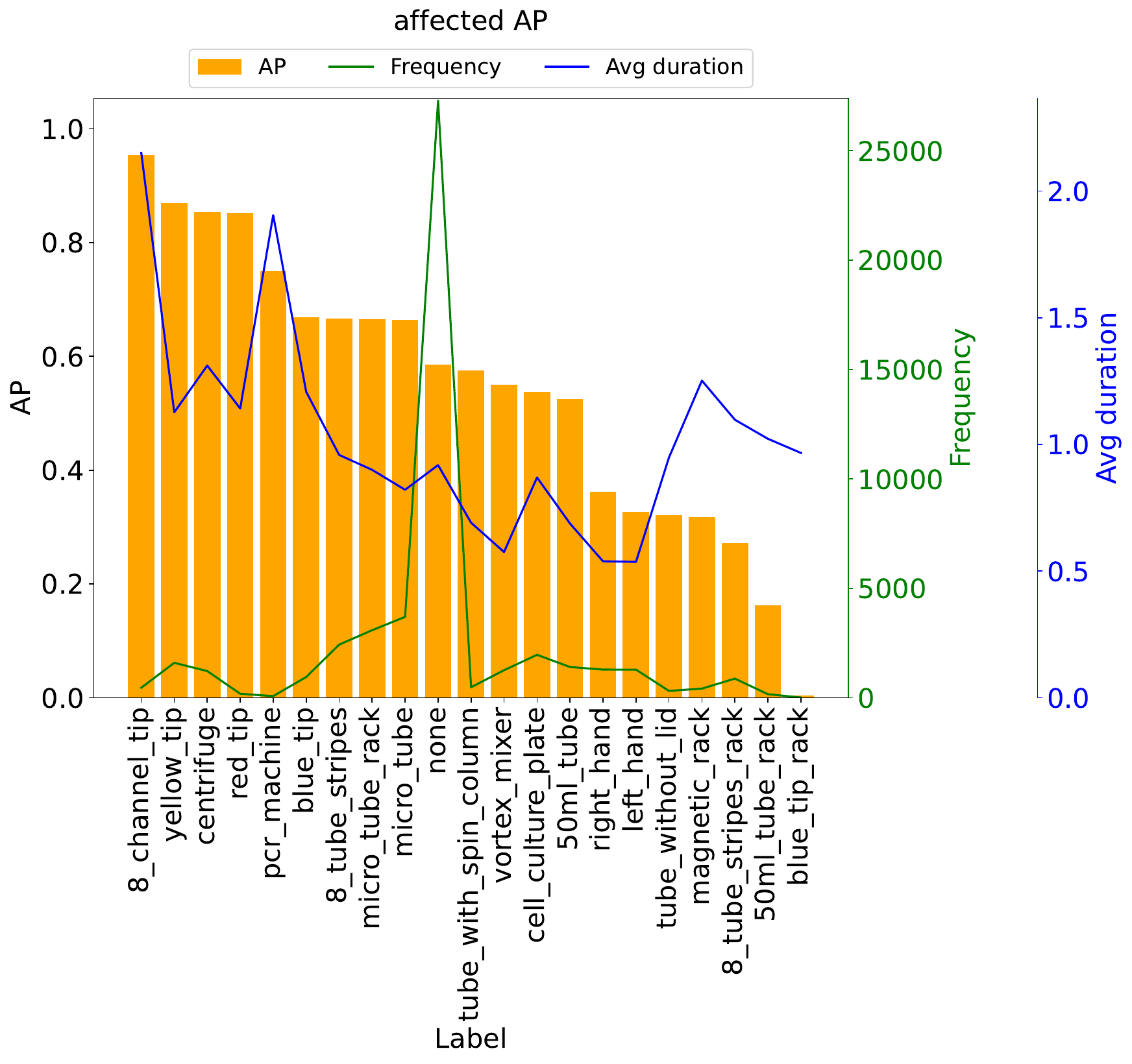}
    \end{minipage}
    \begin{minipage}[htbp]{0.45\linewidth}
        \centering
        \includegraphics[scale=0.16]{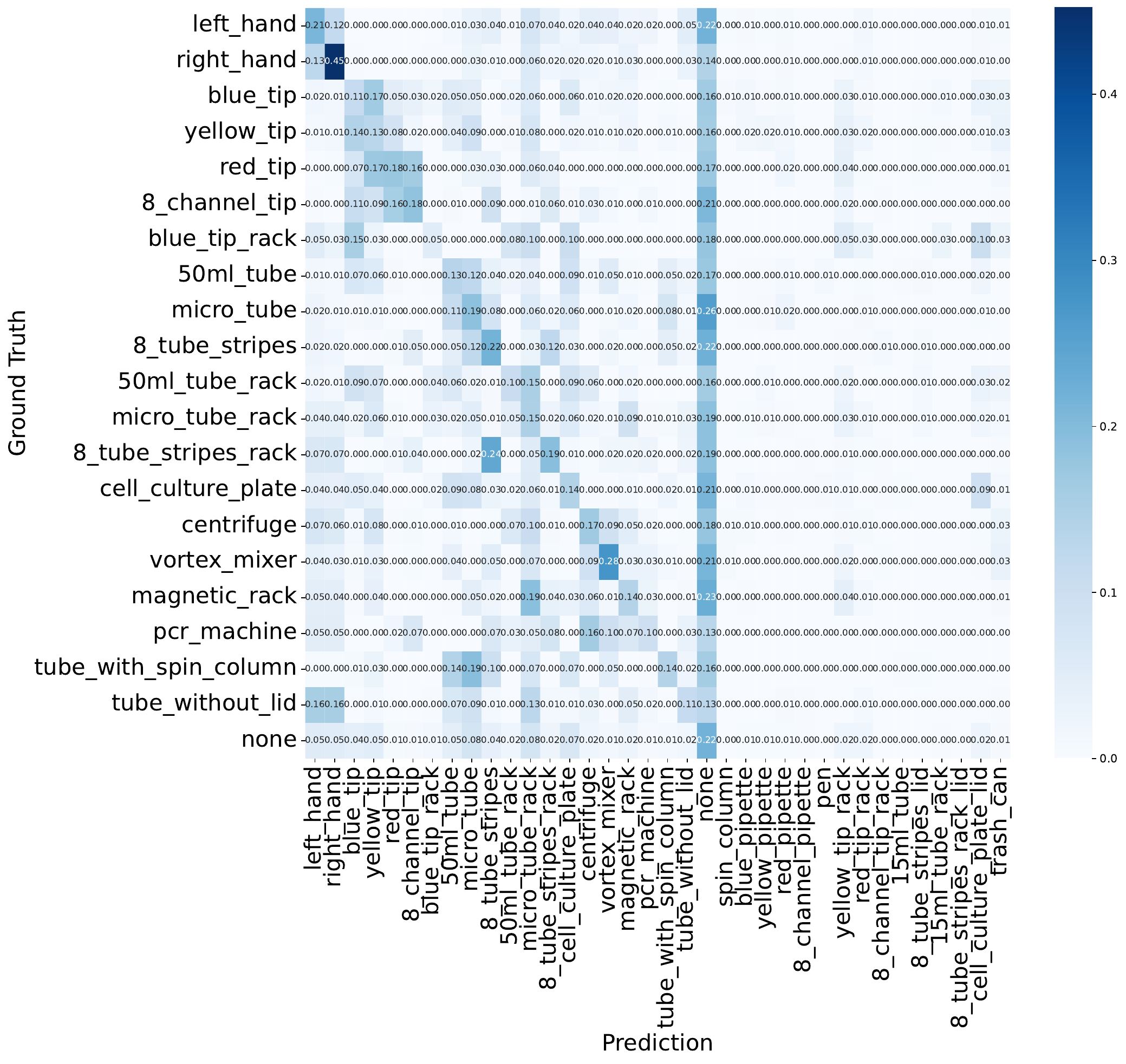}
    \end{minipage}
    \caption{Average Precision (left) and confusion matrix (right) for affected object classes. The green and blue lines in the left figure denote average duration (in seconds) and number of occurrence of each verb.}
    \label{fig:affected_ap_confmat}
\end{figure}

We report class-wise AP and confusion matrix for verb, manipulated object, and affected object in \Fref{fig:verb_ap_confmat}, \Fref{fig:manip_ap_confmat}, and \Fref{fig:affected_ap_confmat}, respectively. Entry in $i$-th row and $j$-th column of a confusion matrix indicates the proportion of the instances predicted as $j$-th label within the instances with $i$-th true label.
All the figures shows the results of the {\bf Separate heads + Set classification} architecture.

The results on verb show the misclassification to more frequent labels such as {\it take} and {\it put}, and confusion between paired operations such as {\it release} and {\it press} or {it take} and {\it put}, because of their symmetric motions.
The results of manipulated and affected objects exhibit confusion between similar-looking objects. (\eg, {\it red tip rack} and {\it yellow tip rack}, {\it red tip} and {\it yellow tip}, {\it tube with spin column} and {\it micro tube}) As for affected objects, we observe the tendency to predict the affected object as {\it none} due to the imbalanced label distribution.

\begin{figure}
    \centering
    \includegraphics[scale=0.5]{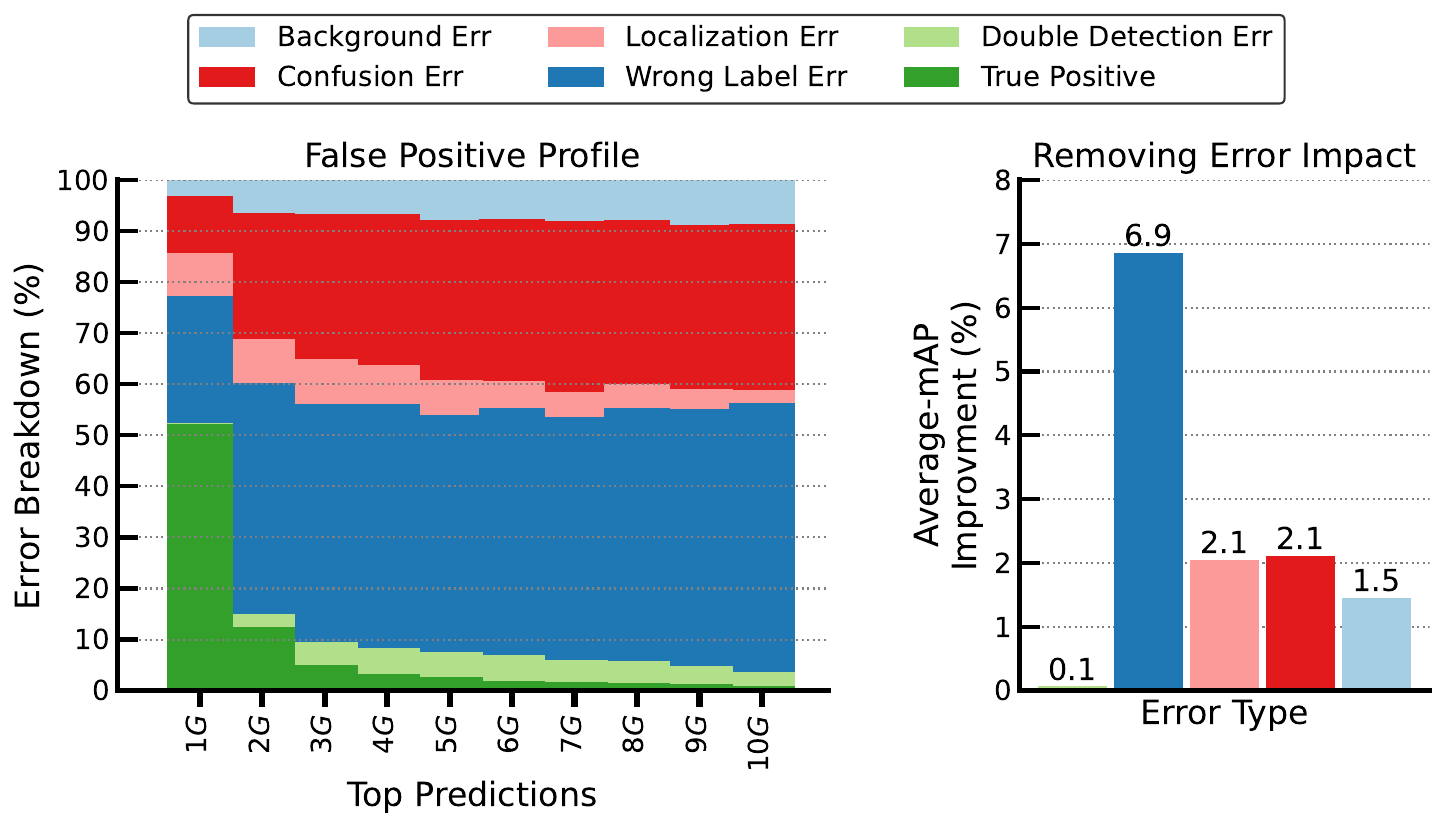}
    \caption{False positive (FP) profiling of the atomic operation prediction results. Left: the error breakdown of five FP error types in top-10G predictions, where G means the number of ground-truth instances. Right: mAP improvement when removing FP instances caused by each error type.}
    \label{fig:action_detection_fp}
\end{figure}

We conduct further analysis following \cite{alwassel2018diagnosing} to analyze false positive/false negative instances in detail. 
\Fref{fig:action_detection_fp} shows the false positive profiling at tIoU of 0.5. The left figure shows the breakdown of five false positive error types in top-10G predictions, where G denotes the number of ground-truth instances. The right displays mAP improvement when removing the false positive instances caused by each error type. The profile indicates that wrong label error is the primary cause of false positive predictions.

\begin{figure}
    \centering
    \includegraphics[scale=0.5]{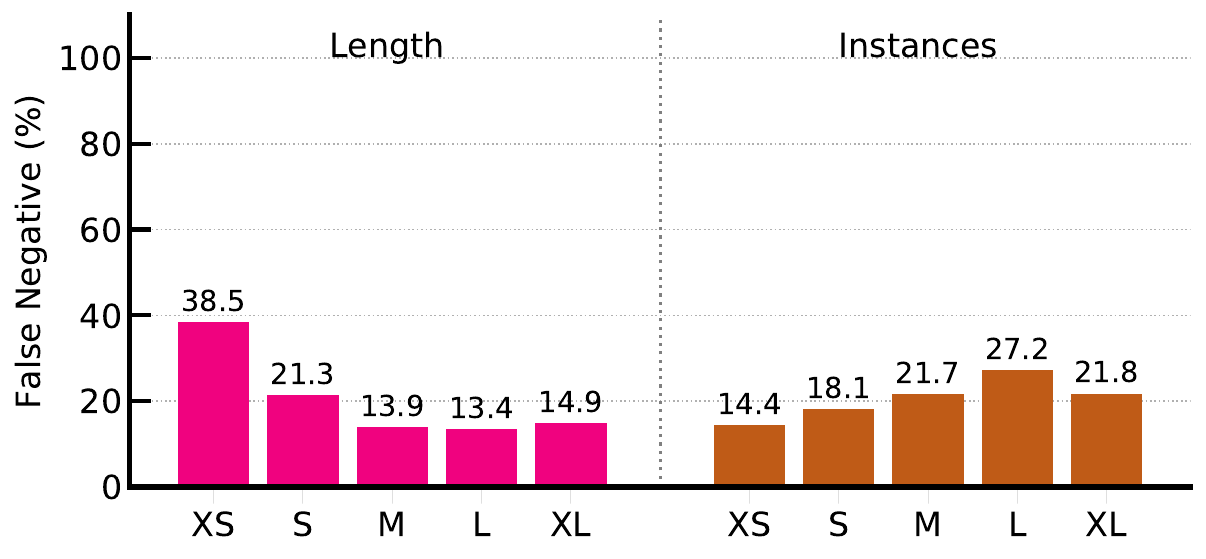}
    \caption{False negative (FN) profiling of our results. This
figure shows the FN rates in groups with different videos features.}
    \label{fig:action_detection_fn}
\end{figure}

\Fref{fig:action_detection_fn} represents the false negative rate under different segment lengths and numbers of instances. We categorize the operation length (in seconds) into five bins: Extra Small (XS): $[0, 0.41]$, Small (S): $(0.41, 0.63]$, Medium (M): $(0.63, 0.87]$, Large (L): $(0.87, 1.23]$, and Extra Long (XL): $(1.23, \infty)$. The number of instances denotes the total number of atomic operations in a video, divided into five groups: Extra Small (XS): $[0, 180]$, Small (S): $(180, 222]$, Medium (M): $(222, 259]$, Large (L): $(259, 296]$, and Extra Long (XL): $(296, \infty)$. The result demonstrates that it is more challenging to detect shorter operations in a video with more atomic operation instances.

\section{Additional Details on Object Detection}
\label{sec:object_detection_details}
\subsection{Implementation Details}
We used (i) two-stage Deformable DETR~\cite{zhu2021deformable} with ResNet-50 backbone and iterative bounding box refinement and (ii) four-scale DINO~\cite{zhang2022dino} with ResNet-50 backbone. We finetuned the models pretrained on COCO~\cite{lin2014microsoft} with the training spit, and evaluate with the test split. We used the models implemented in MMDetection~\cite{chen2019mmdetection} and follow their default training and test settings.

\subsection{Additional Results}
\begin{figure}
    \centering
    \includegraphics[scale=0.4]{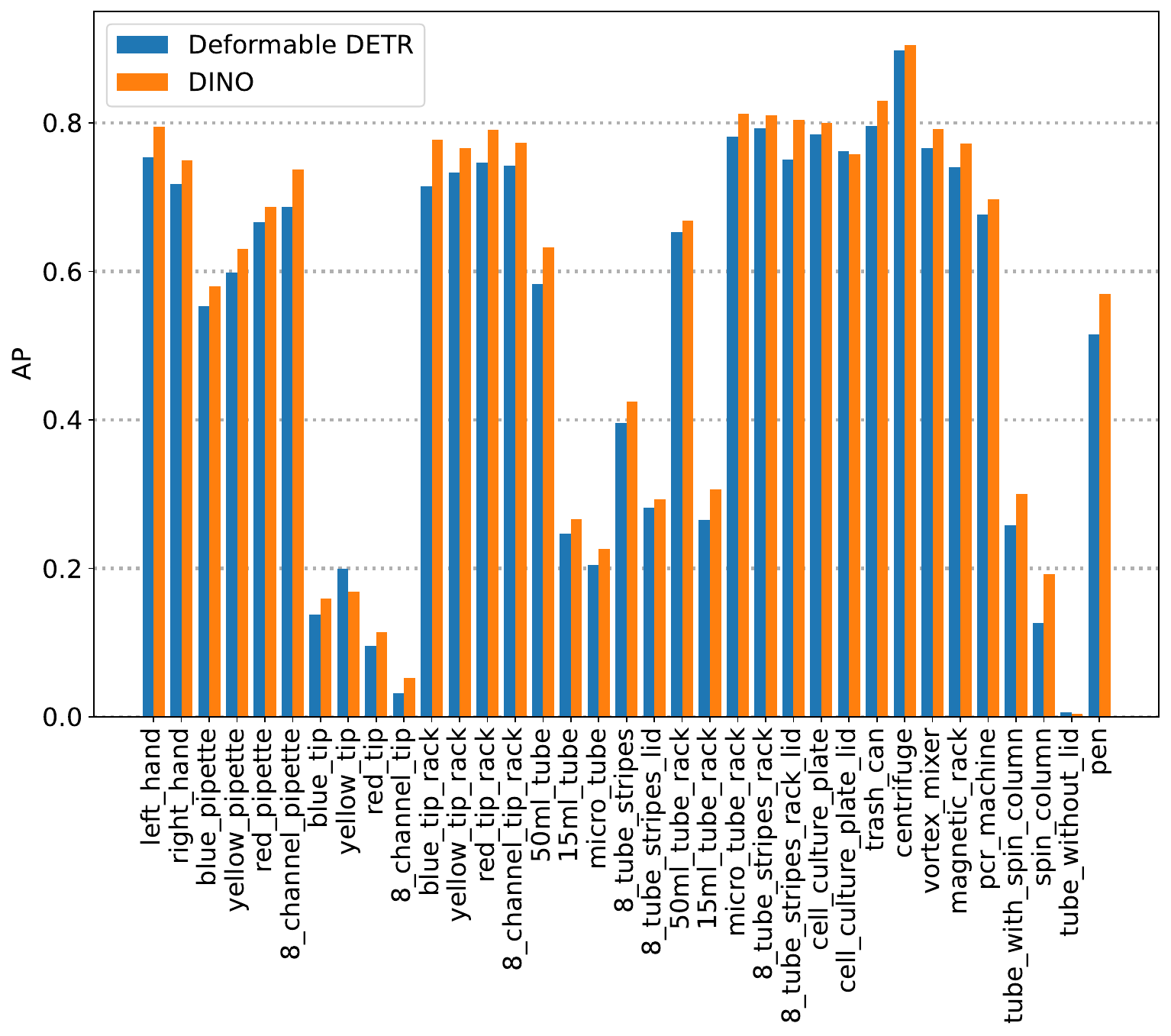}
    \caption{Average Precision (AP) of the detection by Deformable DETR~\cite{zhu2021deformable} and DINO~\cite{zhang2022dino}.}
    \label{fig:object_detection_ap}
\end{figure}

\begin{figure}
    \centering
    \includegraphics[scale=0.21]{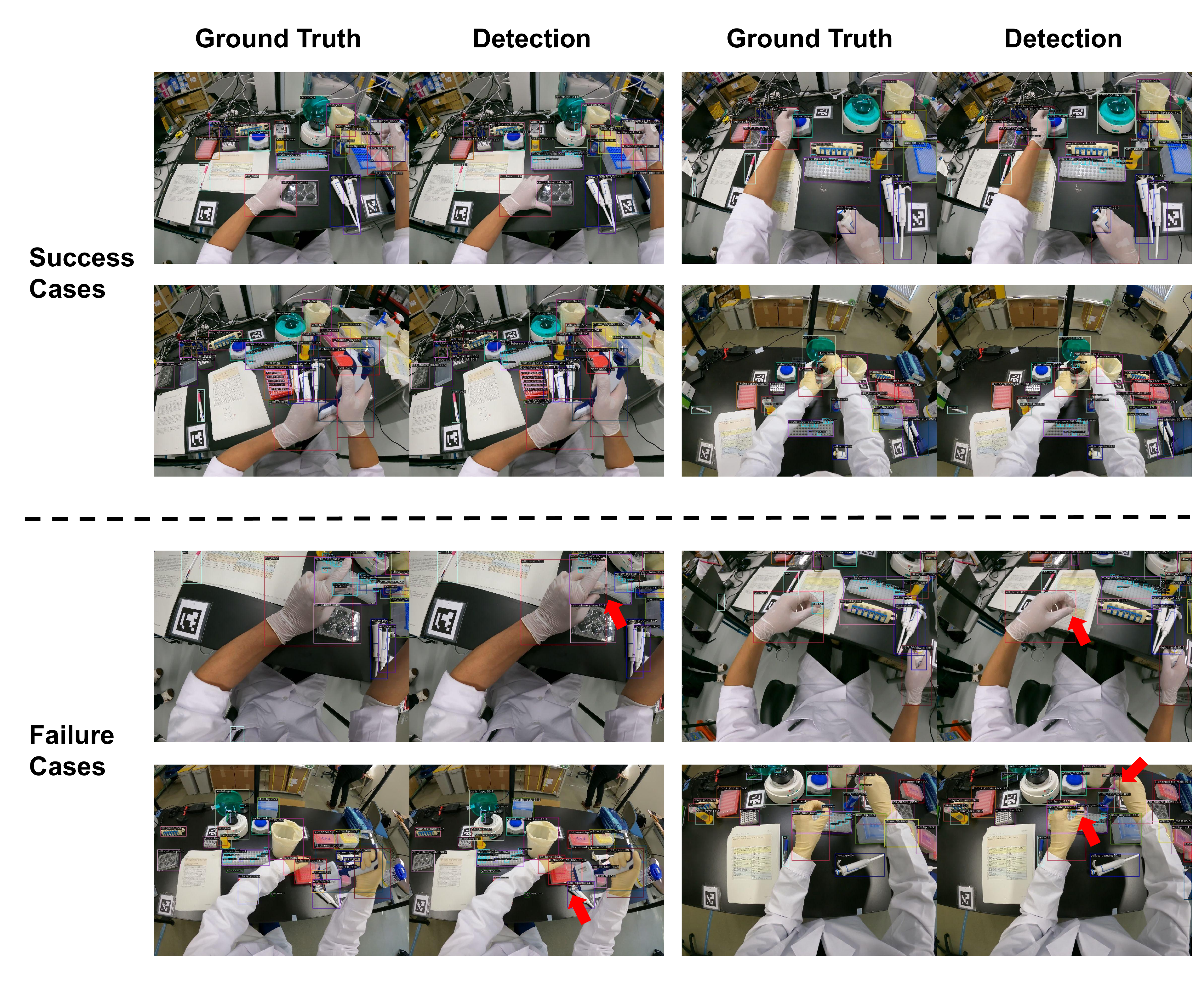}
    \caption{Examples of object detection results. First row shows successful cases and second row displays failure cases. Each example shows correct bounding boxes and labels on left side and predicted ones on right side. Red arrows in failure cases indicate missing detections.}
    \label{fig:object_detection_examples}
\end{figure}

\Fref{fig:object_detection_ap} shows average precision for each object class by the models. While the most of the objects can be detected at over 60 points in AP, small objects such as tips and tubes are hard to detect in both models.
We also show additional qualitative results in \Fref{fig:object_detection_examples}.

\section{Additional Details on Manipulated/Affected Object Detection}
\label{sec:manipulated_affected_object_detection_details}
\subsection{Implementation Details}
We modify the Faster R-CNN~\cite{ren2015faster}-based Hand Object Detector~\cite{shan2020understanding} as described in the main text. We use 4-scale DINO~\cite{zhang2022dino} with a ResNet-50 backbone as an object detector. We first train the object detector for 30 epochs with the default training settings of the original DINO codebase~\cite{zhang2022dino_code}. 
We then train the auxiliary prediction heads with the object detector frozen for 5 epochs. We use SGD optimizer and set an initial learning rate as $10^{-4}$ and batch size as 2. We use the training and test split for training and evaluation, respectively.

\subsection{Analysis and Additional Results}
\begin{figure}
    \centering
    \includegraphics[scale=0.3]{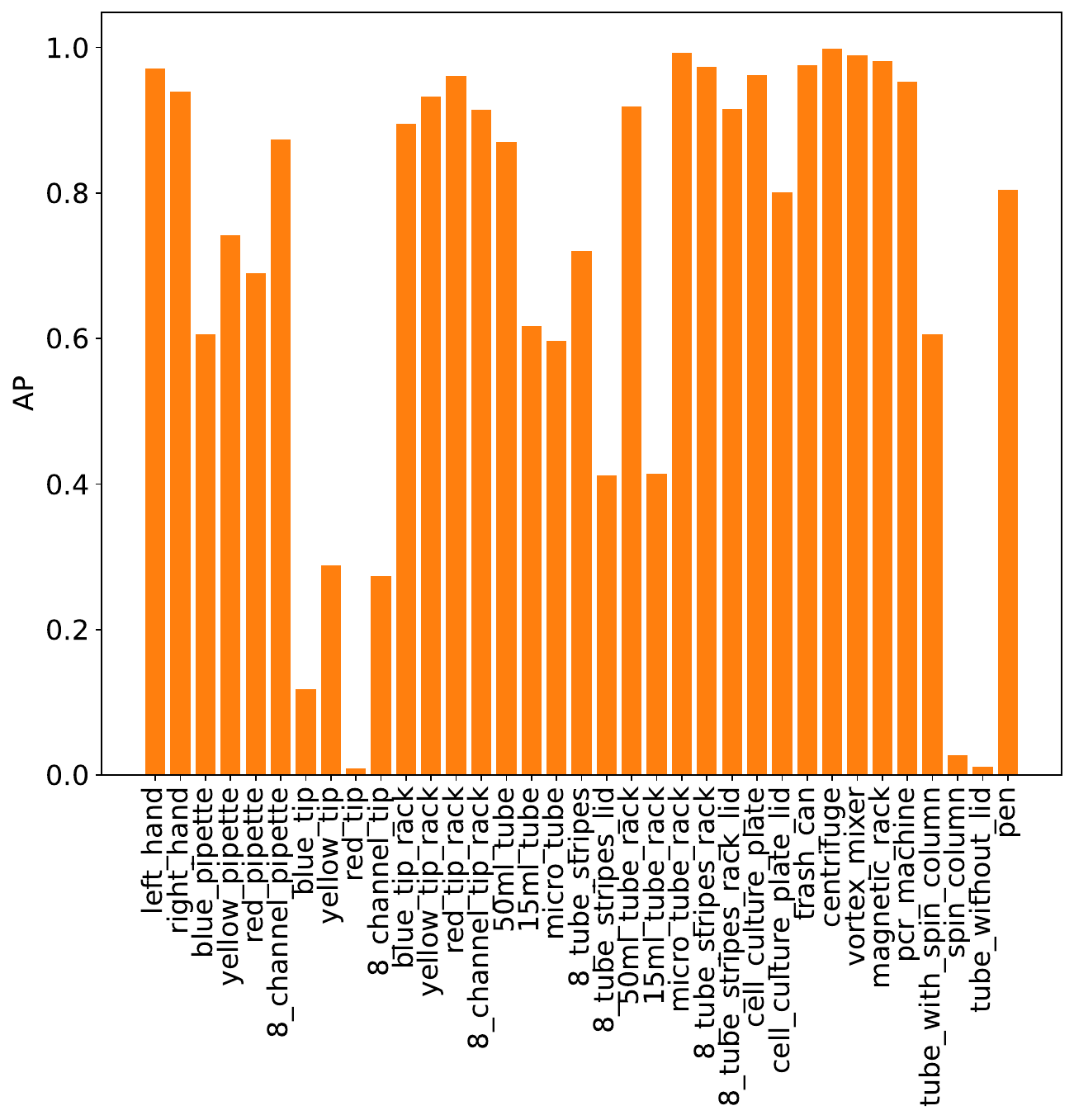}
    \caption{Class-wise $\mathrm{AP}_{50}$ of hand-object detector for manipulated/affected object detection.}
    \label{fig:handobj_ap50}
\end{figure}

\begin{figure}
    \centering
    \includegraphics[scale=0.25]{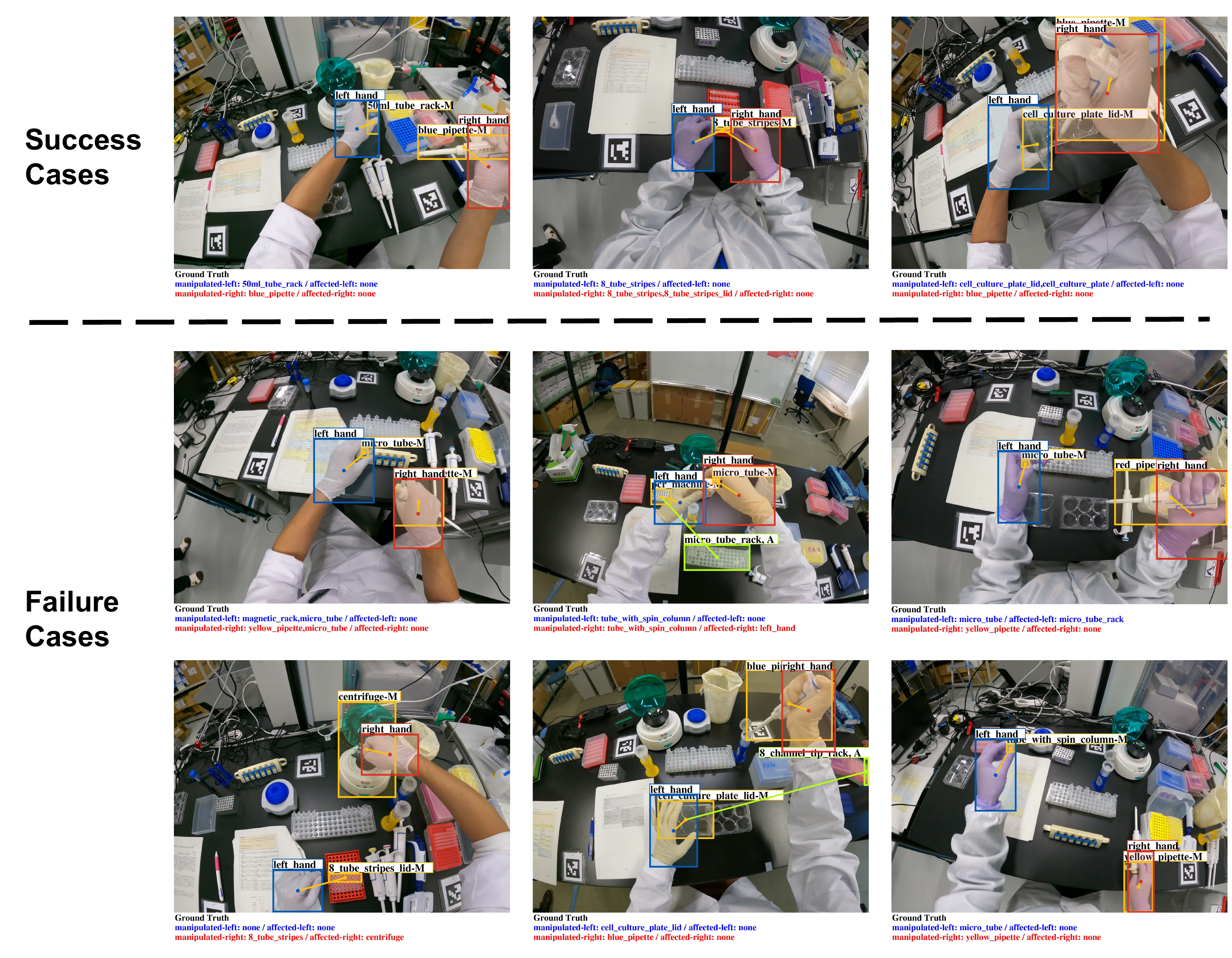}
    \caption{Examples of manipulated/affected object detection results. First row shows successful cases and second and third rows show
    failure cases, including wrong detection of manipulated and affected objects.}
    \label{fig:manipulated_affected_detect_examples}
\end{figure}

We show the base performance of the detector in a standard object detection setting (\Fref{fig:handobj_ap50}).
It presents the $\mathrm{AP}_{50}$ scores for each object class against all the objects in the images, without using the additional heads.
We observe a similar trend as Figure~\Fref{fig:object_detection_ap} that the model struggles in the detection of tips and tubes before determining the manipulated/affected objects, suggesting the need for a stronger backbone object detector for this task.
We also show additional qualitative results in \Fref{fig:manipulated_affected_detect_examples}.

\section{License for FineBio Dataset}
This is a license agreement (referred to herein as the "Agreement") between (licensee's name) (referred to herein as the "Licensee") and National Institute of Advanced Industrial Science and Technology and The University of Tokyo (referred to herein as the "Licensor"). All rights not specifically granted to Licensee in this Agreement are respectively reserved to Licensor.

GRANT OF LICENSE: Licensor hereby grants to Licensee a non-exclusive, non-transferable license to use the Database for the Purpose, without the right to sublicense, pursuant to the terms and conditions of this Agreement. As used in this Agreement, the term "Database" means all or any portion of the data, passwords to decrypt the data, or other intellectual property owned by Licensor and contained in the FineBio database(s) made accessible to Licensee by Licensor pursuant to this Agreement, inclusive of backups, permitted hereunder or subsequently supplied by Licensor. As used in this Agreement, the “Purpose” is limited to: (1) distributing or reproducing any images or videos contained in the Database in a research publication(s), an academic publication(s), or any website through which such publication(s) is made available; (2) using the Database to research, develop, train, evaluate, or improve any software, algorithm, machine learning model, technique, or technology designed to: (a) detect, classify, recognize, retrieve, or understand at least one object, event, place, or activity; (b) model or reconstruct at least one three-dimensional object or environment; and (c) improve any biology application (e.g., laboratory automation); wherein all permitted uses in subsections 2(a)-2(c) of this section may be used for academic research, noncommercial product development, or noncommercial design; and (3) creating and distributing annotations to any images or videos contained in the Database pursuant to the terms and conditions of this Agreement.

INTELLECTUAL PROPERTY: Subject to Licensee’s compliance with the terms and conditions of this Agreement, Licensee retains its intellectual property rights in and to all software, algorithms, machine learning models, annotations, techniques, and technologies developed or otherwise derived from the use of the Database. Such software, algorithms, machine learning models, annotations, techniques, and technologies may be used for academic, commercial, or noncommercial purposes. As between Licensor and Licensee, Licensor retains all intellectual property rights in and to the Database. No other rights or licenses are granted except as expressly provided in this Agreement. All rights not expressly granted under this Agreement are reserved.

PROPRIETARY; COPYRIGHT: Licensee acknowledges that the Database is proprietary to Licensor. Licensee agrees to use the Database only in accordance with the terms of this Agreement. The Database is owned by Licensor and is protected by applicable copyright laws, international treaties, or conventions.

ATTRIBUTION: Any publication or research submitted for publication, academic or otherwise, that is based on, in whole or in part, the Database or use of the Database must include a reference to the following citation in accordance with reasonable academic standards: “FineBio: A Fine-Grained Video Dataset of Biological Experiments with Hierarchical Annotations, Takuma Yagi, Misaki Ohashi, Yifei Huang, Ryosuke Furuta, Shungo Adachi, Toutai Mitsuyama, and Yoichi Sato, 2023.”.

FEEDBACK: Licensor is not obligated to implement any suggestions or feedback Licensee might provide regarding the Database, but to the extent Licensor does so, Licensee is not entitled to any compensation related thereto.

BACKUPS: If Licensee is an organization, it may make that number of copies of the Database necessary for internal use within its organization provided that all information appearing in or on the original labels, including the copyright and trademark notices, is copied onto the labels of the copies.

USES NOT PERMITTED: Except as permitted under the “BACKUPS” section of this Agreement and to make use of the rights provided in the “GRANT OF LICENSE” section of this Agreement, Licensee may not modify or copy the Database. Further, except as permitted herein, all or any portion of the data contained within the Database may not appear in or be visible in any program, dataset, or product, whatsoever, commercial, or otherwise. The licensees must not re-distribute the passwords issued under this Agreement to any third parties. Licensee may not sell, rent, lease, sublicense, lend, time-share, transfer, provide, or provide access to the Database, in whole or in part, to any third party.

ASSIGNMENT: Licensee may not assign this Agreement or any rights herein without the prior written consent of Licensor. Any attempted assignment without such consent shall be null and void.

TERM: The term of the license granted by this Agreement is from Licensee's acceptance of this Agreement by signing this Agreement below until terminated as provided below.

This Agreement automatically terminates without notice if Licensee fails to comply with any provision of this Agreement. Licensee may terminate this Agreement by ceasing use of the Database. Upon any termination of this Agreement, Licensee must destroy any and all copies of the Database. Licensee agrees that all provisions which operate to protect the proprietary rights of Licensor shall remain in force and, as such, survive the term of the Agreement.

DISCLAIMER OF WARRANTIES: THE DATABASE IS PROVIDED "AS-IS" WITHOUT WARRANTY OF ANY KIND INCLUDING ANY WARRANTIES OF PERFORMANCE OR MERCHANTABILITY OR FITNESS FOR A PARTICULAR USE OR PURPOSE OR OF NON-INFRINGEMENT. LICENSEE BEARS ALL RISK RELATING TO QUALITY AND PERFORMANCE OF THE DATABASE.

SUPPORT AND MAINTENANCE: No support, installation, or training by the Licensor is provided as part of this Agreement.

EXCLUSIVE REMEDY AND LIMITATION OF LIABILITY: TO THE MAXIMUM EXTENT PERMITTED UNDER APPLICABLE LAW, LICENSOR SHALL NOT BE LIABLE FOR DIRECT, INDIRECT, SPECIAL, INCIDENTAL, OR CONSEQUENTIAL DAMAGES OR LOST PROFITS RELATED TO LICENSEE’S USE OF OR INABILITY TO USE THE DATABASE, EVEN IF LICENSOR IS ADVISED OF THE POSSIBILITY OF SUCH DAMAGES. LICENSOR SHALL HAVE NO FINANCIAL LIABILITY FOR ANY REASON WHATSOEVER ARISING OUT OF OR RELATING TO THIS AGREEMENT, INCLUDING IN MATTERS RELATED TO THE DATABASE.

EXPORT REGULATION: Licensee agrees to comply with any and all applicable export control laws, regulations, and/or other laws related to embargoes and sanction programs administered by the Office of Foreign Assets Control.

SEVERABILITY: If any provision(s) of this Agreement shall be held to be invalid, illegal, or unenforceable by a court or other tribunal of competent jurisdiction, the validity, legality, and enforceability of the remaining provisions shall not in any way be affected or impaired thereby.

NO IMPLIED WAIVERS: No failure or delay by Licensor in enforcing any right or remedy under this Agreement shall be construed as a waiver of any future or other exercise of such right or remedy by Licensor.

GOVERNING LAW: This Agreement shall be construed and enforced in accordance with the laws of Japan without reference to conflict of laws principles. Licensee hereby consents to the personal jurisdiction of the Tokyo District Court, Japan as the court of the first instance, and waives their right(s) to venue outside of this county.

ENTIRE AGREEMENT: This Agreement constitutes the sole and entire agreement between Licensee and Licensor as to the matters set forth herein and supersedes any previous agreement(s), understanding(s), or arrangement(s) between the parties relating hereto.

BY SIGNING, LICENSEE AGREES TO THE TERMS OF THIS LICENSE AGREEMENT, INTENDING TO BE LEGALLY BOUND HEREBY. IF LICENSEE IS AN ENTITY/ORGANIZATION, THE INDIVIDUAL SIGNING THIS AGREEMENT
 
ON BEHALF OF THE ENTITY/ORGANIZATION REPRESENTS THAT HE/SHE IS AUTHORIZED TO SIGN THIS LICENSE AGREEMENT ON BEHALF OF THE ENTITY/ORGANIZATION. IF LICENSEE DOES NOT AGREE WITH THESE TERMS, LICENSEE MAY NOT USE THE DATABASE.

\section{Datasheet of FineBio Dataset}
\label{sec:datasheet}
We provide a datasheet of FineBio dataset, following datasheets for dataset~\cite{holland2020automation}.

\subsection{Motivation}
\paragraph{For what purpose was the dataset created?}~\\
The dataset was created to visually understanding the activities of people performing biological experiments, potentially targeted for automated experiment recording and laboratory automation.

\paragraph{Who created the dataset (e.g., which team, research group) and on
behalf of which entity (e.g., company, institution, organization)?}~\\
The dataset was created by Artificial Intelligence Research Center, National Institute of Advanced Industrial Science and Technology (AIRC), and Sato Laboratory, Institute of Industrial Science, The University of Tokyo.

\paragraph{Who funded the creation of the dataset?}~\\
This work was funded by JST AIP Acceleration Research JPMJCR20U1.

\paragraph{Any other comments?}~\\
No.

\subsection{Composition}
\paragraph{What do the instances that comprise the dataset represent (e.g.,
documents, photos, people, countries)?}~\\
The instances are videos of people performing mock biological experiments, along with action annotations at multiple levels (protocol, step, atomic operation, and object location). Participants are researchers or workers who have/had experience in conducting biological experiments.

\paragraph{How many instances are there in total (of each type, if appropriate)?}~\\
There are 226 multi-view videos in total from 32 participants (16 male, 16 female), and three optional videos that contain major procedural errors, removed from evaluation.

\paragraph{Does the dataset contain all possible instances or is it a sample (not
necessarily random) of instances from a larger set?}~\\
We intended to collect various tool manipulation behaviors from people who have experiences in performing basic biological experiments.

\paragraph{What data does each instance consist of?}~\\
One participant collected 5-10 video trials from seven protocols.
Each video set consists of five third-person videos taken from a fixed camera installed on a platform and one first-person video taken from a head-mounted camera.
For each trial, action, and object annotations on step (instructions described in protocols), atomic operation (minimum meaningful actions consisted of a verb, manipulated object, and affected object (optional)), object location, and object states (whether in contact, and whether the object is manipulated, affected, or not).

\paragraph{Is there a label or target associated with each instance?}~\\
The step labels are the instruction texts taken from the seven protocols.
Some step labels are shared among protocols.
The verb labels consist of ten verbs (put, take, press, release, insert, detach, open, close, eject, shake).
The manipulated/affected objects labels consist of two hand labels (right/left hand) and 33 types of equipment that appeared in the experiments.

\paragraph{Is any information missing from individual instances?}~\\
No data is missing.

\paragraph{Are relationships between individual instances made explicit (e.g.,
users’ movie ratings, social network links)?}~\\
Videos are associated with unique and anonymized participant ids.

\paragraph{Are there recommended data splits (e.g., training, development/
validation, testing)?}~\\
Yes. We provide training/validation/testing splits divided by participants. Refer to Section~\ref{ssec:release} for details.

\paragraph{Are there any errors, sources of noise, or redundancies in the
dataset?}~\\
Although we took great care to ensure that the participants followed the procedures described in the protocol, a few videos contain step-level errors such as skipping a step or switching the order of steps, and three of them were removed from evaluation. See Section~\ref{ssec:annotation_details} for details.

\paragraph{Is the dataset self-contained, or does it link to or otherwise rely on
external resources (e.g., websites, tweets, other datasets)?}~\\
Yes, the dataset is entirely self-contained.

\paragraph{Does the dataset contain data that might be considered confidential
(e.g., data that is protected by legal privilege or by doctor–patient
confidentiality, data that includes the content of individuals’ nonpublic
communications)?}~\\
No. However, we intentionally redacted the audio information because it is less useful for the purpose of this dataset, and may include private communications.

\paragraph{Does the dataset contain data that, if viewed directly, might be offensive,
insulting, threatening, or might otherwise cause anxiety?}~\\
No.

\paragraph{Does the dataset identify any subpopulations (e.g., by age, gender)?}~\\
Due to the recording location, all the participants are Asian (Japanese).

\paragraph{Is it possible to identify individuals (i.e., one or more natural persons),
either directly or indirectly (i.e., in combination with other
data) from the dataset?}~\\
Yes in theory, because the videos are not anonymized for naturalness and the face wearing a mask of the participants is visible.
However, the risk is minimal and the likelihood of identifying an individual from an image is very low.
We took consent to the participants to make their appearance public to approved researchers.

\paragraph{Does the dataset contain data that might be considered sensitive in
any way (e.g., data that reveals race or ethnic origins, sexual orientations,
religious beliefs, political opinions or union memberships, or
locations; financial or health data; biometric or genetic data; forms of
government identification, such as social security numbers; criminal
history)?}~\\
No.

\paragraph{Any other comments?}~\\
No.

\subsection{Collection Process}
\paragraph{How was the data associated with each instance acquired?}~\\
The video data was newly collected by video recording.

\paragraph{What mechanisms or procedures were used to collect the data (e.g.,
hardware apparatuses or sensors, manual human curation, software programs, software APIs)?}~\\
We used six GoPro HERO9 cameras to record the videos.

\paragraph{If the dataset is a sample from a larger set, what was the sampling
strategy (e.g., deterministic, probabilistic with specific sampling
probabilities)?}~\\
N/A.

\paragraph{Who was involved in the data collection process (e.g., students,
crowdworkers, contractors) and how were they compensated (e.g.,
how much were crowdworkers paid)?}~\\
Three authors (TY, MO, and TM) was involved in the data collection process as a supervisor.
Participants are researchers or workers who have/had experiences in conducting biological experiments.
They participated in the recording as part of their job duties.

\paragraph{Over what timeframe was the data collected?}~\\
All data were collected from October to December 2022.

\paragraph{Were any ethical review processes conducted (e.g., by an institutional
review board)?}~\\
Yes, the experimental plan was reviewed and approved by the committee for ergonomic experiments established within National Institute of Advanced Industrial Science and Technology.
We have attached the (translated version of) approval from the committee for ergonomics experiments, the participant instruction document, and the consent form in the supplementary materials.

\paragraph{Did you collect the data from the individuals in question directly, or
obtain it via third parties or other sources (e.g., websites)?}~\\
We met with the individuals and recorded their videos in-person.

\paragraph{Were the individuals in question notified about the data collection?}~\\
Yes. Following the ethical review process, the participants were informed of the purpose of the experiment, the content of the data to be collected, its management, and the extent to which it will be made public.

\paragraph{Did the individuals in question consent to the collection and use of
their data?}~\\
Yes.

\paragraph{If consent was obtained, were the consenting individuals provided
with a mechanism to revoke their consent in the future or for certain
uses?}~\\
Yes. They have the right to revoke the consent by letter no later than 30 days after participation.

\paragraph{Has an analysis of the potential impact of the dataset and its use
on data subjects (e.g., a data protection impact analysis) been conducted?}~\\
Yes. The risks on data publication was reviewed and approved by the committee for ergonomic experiments.

\paragraph{Any other comments?}~\\
None.

\subsection{Preprocessing/cleaning/labeling}
\paragraph{Was any preprocessing/cleaning/labeling of the data done (e.g., discretization or bucketing, tokenization, part-of-speech tagging, SIFT
feature extraction, removal of instances, processing of missing values)?}~\\
We retook the experiments when the participant made an apparent mistake or violated the instructions written in the protocols.
We removed the recordings where one or more videos were missing due to hardware issues.
All the multi-view videos are temporally synchronized, and the unnecessary segments are discarded in the release.
Audio information are removed from all the videos since it is less necessary and may include confidential information.

\paragraph{Was the ``raw'' data saved in addition to the preprocessed/cleaned/labeled data (e.g., to support unanticipated future uses)?}~\\
No, we do not plan to release the raw data because the processed data includes sufficient information for the intended purposes.

\paragraph{Is the software that was used to preprocess/clean/label the data available?}~\\
No. The preprocessing process was mostly done by manual inspection.

\paragraph{Any other comments?}~\\
None.

\subsection{Uses}
\paragraph{Has the dataset been used for any tasks already?}~\\
No.

\paragraph{Is there a repository that links to any or all papers or systems that
use the dataset?}~\\
N/A.

\paragraph{What (other) tasks could the dataset be used for?}~\\
The dataset could be used for various visual recognition tasks such as action recognition/anticipation, hand-object interaction modeling, and developing fundamental models in laboratory automation and automatic recording of biological experiments.

\paragraph{Is there anything about the composition of the dataset or the way it was collected and preprocessed/cleaned/labeled that might impact future uses?}~\\
There are very low risks of harming specific individuals or harmful uses because it does not include any confidencial information.

\paragraph{Are there tasks for which the dataset should not be used?}~\\
For convenience and safety, we recorded ``mock'' experiments which are taken from real experiments but do not use real materials and reagents.
Therefore, the recognition model trained by this dataset cannot be directly applied to actual biological experiments for the purpose of automated experiment recording.

\paragraph{Any other comments?}~\\
None.

\subsection{Distribution}
\paragraph{Will the dataset be distributed to third parties outside of the entity (e.g., company, institution, organization) on behalf of which the dataset was created?}~\\
Yes, the dataset will be distributed to approved researchers for non-commercial research purposes, to facilitate new research in computer vision and laboratory automation.

\paragraph{How will the dataset will be distributed (e.g., tarball on website, API,
GitHub)?}~\\
The documentation and the evaluation code are publicly distributed under the official repository of the Artificial Intelligence Research Center, National Institute of Advanced Industrial Science and Technology (\url{https://github.com/aistairc/FineBio}).
The videos will be distributed as a password-protected zip file via ABCI cloud storage (\url{https://docs.abci.ai/en/abci-cloudstorage/}).

To access the dataset, the user must agree to the license agreement (\url{https://finebio.s3.abci.ai/FineBio_License_Agreement.pdf}).
The user is asked to submit the signed license agreement via the dataset request form (see attached file for details), and the application will be reviewed by the authors.
Once approved, an e-mail providing the URL and the password will be sent to the user's contact e-mail address.

\paragraph{When will the dataset be distributed?}~\\
The dataset distribution will starts in February 2024.

\paragraph{Will the dataset be distributed under a copyright or other intellectual property (IP) license, and/or under applicable terms of use (ToU)?}~\\
The dataset will be distributed under the CC BY-NC 4.0 license (\url{https://creativecommons.org/licenses/by-nc/4.0/deed.en}).
In addition to the above license, there is a request to cite this paper after they are published to some venues.

\paragraph{Have any third parties imposed IP-based or other restrictions on the data associated with the instances?}~\\
No.

\paragraph{Do any export controls or other regulatory restrictions apply to the dataset or to individual instances?}~\\
No.

\paragraph{Any other comments?}~\\
None.

\subsection{Maintenance}
\paragraph{Who will be supporting/hosting/maintaining the dataset?}~\\
Takuma Yagi and Toutai Mitsuyama will support the dataset.
The dataset will be hosted to the ABCI cloud storage that is maintained by National Institute of Advanced Industrial Science and Technology.

\paragraph{How can the owner/curator/manager of the dataset be contacted (e.g., email address)?}~\\
Please contact by e-mails ( TY: \texttt{takuma.yagi@aist.go.jp}, TM: \texttt{mituyama-toutai@aist.go.jp} ).

\paragraph{Is there an erratum?}~\\
N/A since this is the first release.

\paragraph{Will the dataset be updated (e.g., to correct labeling errors, add new instances, delete instances}~\\
Updates will be posted on the dataset (GitHub) webpage.

\paragraph{If the dataset relates to people, are there applicable limits on the retention of the data associated with the instances (e.g., were the individuals in question told that their data would be retained for a fixed period of time and then deleted)?}~\\
No, we did not set a specific limit.

\paragraph{Will older versions of the dataset continue to be supported/hosted/maintained?}~\\
Yes, we plan to keep older versions in the original website if updated.

\paragraph{If others want to extend/augment/build on/contribute to the dataset, is there a mechanism for them to do so?}~\\
External people can make a proposal through the issues in the GitHub page.

\paragraph{Any other comments?}~\\
None.

\end{document}